\documentclass{article} 

\usepackage{amsmath,amsfonts,bm}

\def\eqref#1{equation~\ref{#1}}

\def\1{\bm{1}}

\def\w{{\bf w}}

\def\x{{\bf x}}

\def\y{{\bf y}}
\def\I{{\bf I}}

\DeclareMathAlphabet{\mathsfit}{\encodingdefault}{\sfdefault}{m}{sl}
\SetMathAlphabet{\mathsfit}{bold}{\encodingdefault}{\sfdefault}{bx}{n}

\usepackage{fullpage}
\usepackage{hyperref}
\hypersetup{colorlinks,linkcolor={cvprblue},citecolor={cvprblue},urlcolor={RoyalBlue}} 
\usepackage{url}
\usepackage{graphicx}
\usepackage{cleveref}
\usepackage{booktabs}       
\usepackage{xcolor}         
\usepackage{float}
\usepackage{colortbl}
\usepackage{enumitem}
\usepackage{multirow}
\usepackage[mathscr]{eucal}
\usepackage[export]{adjustbox}
\usepackage{multicol,lipsum}
\usepackage{arydshln}
\usepackage{pifont}
\usepackage[table, dvipsnames, svgnames]{xcolor}
\usepackage{comment}
\usepackage{wrapfig}
\usepackage{subcaption}
\usepackage{makecell}
\usepackage{pifont}
\usepackage[numbers]{natbib}
\usepackage{marvosym}

\definecolor{cvprblue}{rgb}{0.21,0.49,0.74}
\definecolor{purple}{rgb}{0.56,0.27,0.68}
\definecolor{red}{rgb}{0.95,0.4,0.4}
\definecolor{purered}{rgb}{1,0,0}
\definecolor{blue}{rgb}{0.4,0.4,0.95}
\definecolor{darkblue}{rgb}{0,0,0.8}
\definecolor{lightblue}{rgb}{127,153,240}
\definecolor{grey}{rgb}{0.6,0.6,0.6}
\definecolor{col1}{RGB}{232, 161, 148}
\definecolor{col11}{RGB}{255, 228, 228}
\definecolor{col2}{RGB}{148, 187, 232}
\definecolor{col33}{RGB}{206, 239, 255}
\definecolor{col3}{RGB}{225, 239, 220} 
\definecolor{lightgrey}{rgb}{0.85,0.85,0.85}
\definecolor{lightyellow}{RGB}{255,195,78}
\definecolor{lightlightgrey}{rgb}{0.9,0.9,0.9}
\definecolor{verylightBG}{rgb}{0.9,0.99,0.99}
\definecolor{darkgreen}{rgb}{0., 0.85, 0.5}
\definecolor{gtred}{RGB}{204, 0, 0}
\definecolor{predgreen}{RGB}{31, 237, 31}
\definecolor{figGreen}{RGB}{56, 118, 29}

\newtheorem{definition}{Definition}
\newtheorem{theorem}{Theorem}
\newtheorem{corollary}{Corollary}

\definecolor{colorNoSelection}{RGB}{227, 155, 0}
\newcommand{\colorNoSelection}[1]{\textcolor{colorNoSelection}{#1}}
\newcommand{\NoSelection}{\colorNoSelection{$\mathbf{\bullet}$\,}}

\definecolor{gFONE}{RGB}{154, 113, 196}
\newcommand{\colorgFONE}[1]{\textcolor{gFONE}{#1}}
\newcommand{\gFONE}{\colorgFONE{$\mathbf{\bullet}$\,}}

\definecolor{IDtrn}{RGB}{140, 85, 74}
\newcommand{\colorIDtrn}[1]{\textcolor{IDtrn}{#1}}
\newcommand{\IDtrn}{\colorIDtrn{$\mathbf{\bullet}$\,}}

\definecolor{Rtr}{RGB}{44, 160, 44}
\newcommand{\colorRtr}[1]{\textcolor{Rtr}{#1}}
\newcommand{\Rtr}{\colorRtr{$\mathbf{\bullet}$\,}}

\definecolor{ArithmeticMean}{RGB}{188, 189, 33}
\newcommand{\colorArithmeticMean}[1]{\textcolor{ArithmeticMean}{#1}}
\newcommand{\ArithmeticMean}{\colorArithmeticMean{$\mathbf{\bullet}$\,}}

\definecolor{BordaCount}{RGB}{90, 169, 162}
\newcommand{\colorBordaCount}[1]{\textcolor{BordaCount}{#1}}
\newcommand{\BordaCount}{\colorBordaCount{$\mathbf{\bullet}$\,}}

\definecolor{FTWO}{RGB}{231, 110, 141}
\newcommand{\colorFTWO}[1]{\textcolor{FTWO}{#1}}
\newcommand{\FTWO}{\colorFTWO{$\mathbf{\bullet}$\,}}

\definecolor{FHALF}{RGB}{231, 144, 141}
\newcommand{\colorFHALF}[1]{\textcolor{FHALF}{#1}}
\newcommand{\FHALF}{\colorFHALF{$\mathbf{\bullet}$\,}}

\let\citeoriginal\cite
\renewcommand{\cite}[1]{{\normalsize\citeoriginal{#1}}}

\title{Enabling Validation for Robust Few-Shot Recognition}

\author{Hanxin Wang$^{1, }$\thanks{The first two authors make equal contributions. $^\text{\Letter}$Corresponding author.} \quad Tian Liu$^{2, *}$ \quad Shu Kong$^{1, 3, \text{\Letter}}$ \\
$^1$University of Macau \quad $^2$Texas A\&M University \quad $^3$IAIBS \\
{\small \em website: \url{https://hannawang09.github.io/projects/vest/}}
\vspace{-3mm}
}

\begin{document}

\date{}
\maketitle

\begin{abstract}
A key challenge in Few-Shot Recognition (FSR) is the scarcity of labeled task-specific data for training.
Contemporary methods mitigate this challenge by exploiting pretrained Vision-Language Models (VLMs) for their learned transferable features.
Some recent ones augment the scarce data by retrieving relevant yet out-of-distribution (OOD) examples from external data sources.
However,
they largely overlook an equally fundamental challenge: the absence of validation data for model selection and hyperparameter tuning. 
As a result, the adapted VLMs overfit the few-shot training data and generalize poorly to OOD test data.
We show that retrieved data provide a natural source of validation, but their direct use creates a fundamental paradox. 
Since the retrieved samples are inherently OOD w.r.t the few-shot ID training data, adapting a VLM to the ID data constantly degrades its performance on the retrieved data. 
Consequently, conventional validation relying on the accuracy of retrieved data favors no adaptation and prevents improved generalization.
To resolve this dilemma, we formulate validation as a multi-objective optimization (MOO) problem between performance gains on the few-shot ID data and the performance preservation on the retrieved data.
We introduce a practically effective and theoretically grounded validation strategy that selects an operating point on the Pareto front.
This strategy enables hyperparameter tuning, 
which had been unrealistically done by using test data,
thereby mitigating overfitting and improving generalization. 
We integrate it into a stage-wise finetuning pipeline,
termed \textit{Validation-Enabled Stage-wise Tuning (VEST)}.
Extensive experiments on five established benchmarks show that VEST consistently outperforms existing approaches on both ID and OOD test data.

\end{abstract}

\begin{figure}[h]
    \begin{minipage}{.48\textwidth}
        \centering
        \vspace{-4.2mm}
        \includegraphics[width=0.9\linewidth, clip=true,trim = 0mm 0mm 0mm 0mm]{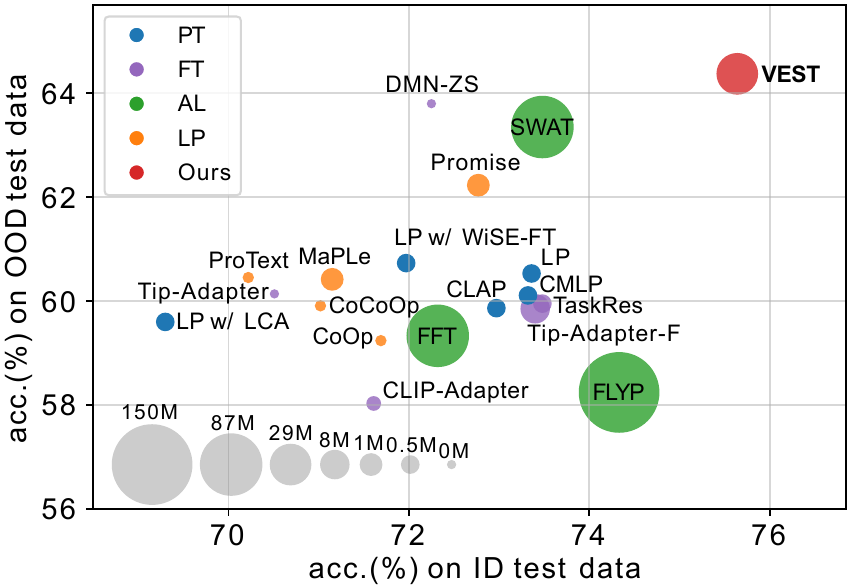}
        \vspace{-2mm}
    \end{minipage}
    \hfill
    \begin{minipage}{.51\textwidth}
        \caption{\small 
        \textbf{Summary of benchmarking results of VLM adaptation methods} 
        w.r.t accuracies on the ImageNet ID and OOD test data.
        We use circle size to mark the number of learned parameters of each method.
        All the compared methods adapt a pretrained CLIP ViT-B/16 VLM~\citep{radford2021learning} with the 16-shot ImageNet ID data.
        Refer to  \S\ref{sec:exp} for dataset details, Fig.~\ref{fig:vlm_adapt} for the compared methods, and \Cref{tab:baseline} for detailed results.
        Our method {\bf VEST} (\S\ref{ssec:ours}) significantly outperforms existing methods in both ID and OOD accuracies, with a moderate number of learned parameters.
        Notably, unlike VEST, CLAP, FFT, and SWAT, other methods artificially use the ImageNet ID test set for tuning.
        }
        \label{fig:sota_ood}
    \end{minipage}
    \vspace{-7mm}
\end{figure}

\section{Introduction}
\label{sec:intro}

Few-Shot Recognition (FSR) formulates a classification task that provides a small amount of labeled training data for object categories of interest. 
Recent FSR methods mitigate this data scarcity issue by adapting pretrained Vision-Language Models (VLMs)~\citep{radford2021learning, li2021align} for their learned transferable features.
In real test-time scenarios, 
the FSR models inevitably encounter out-of-distribution (OOD) observations that distributionally deviate from the task-specific in-distribution (ID) training data.
This challenges the generalization of the FSR models,
motivating our work on \emph{robust FSR} with VLM adaptation.

{\bf Status Quo.}
Current FSR methods commonly leverage pretrained VLMs, adapting them via parameter-efficient finetuning (PEFT) such as prompt tuning (PT)~\citep{zhou2022learning,  chen2022plot, khattak2023maple, roy2024consistencyguided, Khattak2024ProText}, 
linear probing (LP)~\citep{clap24, lin2023multimodality, shi2024lca}, and adapter learning (AL)~\citep{zhang2022tip, gao2023clip}.
The recent work~\citep{liu2025few} motivates FSR from a data-annotation perspective, prioritizing accuracy over parameter efficiency, and shows that finetuning the entire visual encoder significantly outperforms PEFT methods w.r.t classification accuracy.
It also recommends leveraging external data sources, e.g., the VLM's pretraining data, to augment the few-shot training examples.
Moreover,
while early FSR methods artificially tune their models on test sets,
the current literature converges to a more realistic setting that admits the absence of validation data~\citep{lin2023multimodality, clap24, liu2025few, liu2026solving}.
Nevertheless,
a crucial aspect remains unaddressed: 
the robustness of FSR methods to OOD test data, whose distribution deviates from the training data.
This issue is exacerbated by a complete lack of validation data.

{\bf  Insights.}
The core challenge of FSR is data scarcity, extending beyond limited training data to a complete absence of validation data.
The recent work~\citep{liu2025few} retrieves task-relevant yet noisy samples from external data sources to augment the few-shot training data.
It defaults hyperparameters (e.g., epochs and learning rate) to those widely used in the literature.
Yet, no prior work has exploited the retrieved data for validation.
The retrieved samples are inherently OOD compared with the task-specific ID training data.
Consequently, finetuned VLM yields degraded performance on the retrieved samples and directly using them as validation data for checkpoint selection selects the original VLM without any finetuning, 
restricting the generalization to both ID and OOD test data (Fig.~\ref{fig:dilemma}).
We address this dilemma with a novel validation strategy,
which harmonizes the performance gain and preservation on the ID training data and the retrieved data, respectively (Fig.~\ref{fig:validation}).
Our validation enables selecting a checkpoint that generalizes to both ID and OOD test data.
Importantly, it also allows tuning hyperparameters, such as pretrained layers to finetune, towards better generalization (Fig.~\ref{fig:layer-selection}).
We incorporate the strategy and derive a simple yet effective finetuning pipeline:
\textbf{V}alidation-\textbf{E}nabled \textbf{S}tage-wise \textbf{T}uning (\textbf{VEST}).
Extensive experiments demonstrate that VEST significantly outperforms existing VLM adaptation methods for FSR (Fig.~\ref{fig:sota_ood}).

{\bf Contributions.}
We make three key contributions:
\begin{enumerate}[leftmargin=15pt, topsep=0pt, itemsep=5pt,parsep=-2pt]
    \item We establish the robust FSR problem and comprehensively evaluate a variety of recent methods.

    \item We propose a novel validation strategy by exploiting the few-shot ID training data and external data, enabling checkpoint selection and hyperparameter tuning to improve generalization.

    \item We present a robust finetuning pipeline with our validation strategy, significantly outperforming existing VLM-based FSR methods w.r.t. both ID and OOD accuracy.
    
\end{enumerate}

\section{Related work}
\label{sec:related-work}

{\bf Few-Shot Recognition (FSR).}
Contemporary methods exploit a pretrained VLM for its learned transferable features.
Most of them freeze the VLM and only learn prompt tokens~\citep{zhou2022learning, zhou2022conditional, yao2023visual, khattak2023maple, roy2024consistencyguided} or an adapter~\citep{gao2023clip, zhang2022tip, song2023meta, zhang2024dual}.
Some finetune the VLM~\citep{liu2026solving, liu2025few, goyal2023finetune, kumar2022finetuning, madan2024revisiting, liu2026visual, ma2026autoexpert} but \citep{kumar2022finetuning} shows that finetuning can destroy pretrained features and degrade the generalization ability.
\citep{wortsman2022robust} mitigates this issue by ensembling the finetuned and the pretrained models.
However, no prior work has comprehensively benchmarked VLM adaptation methods w.r.t  OOD generalization, measuring accuracies on both ID and OOD test data.
Thus, improving the OOD generalization of adapted VLM with limited ID training data remains an open problem. 
Moreover, before recent work \citep{lin2023multimodality, clap24, liu2025few},  many prior FSR methods tuned hyperparameters on a held-out labeled validation set or even the test set.
Such an artificial practice causes data leakage and produces over-optimistic results and unfair comparisons.
In this work, we adopt a realistic setup that eschews a validation set~\citep{clap24, liu2025few} and conduct the first comprehensive study of VLM adaptation for robust FSR.

{\bf Open Data.}
The abundance of publicly available data not only facilitates training more generalizable and robust models~\citep{kong2021opengan, hendrycks2018deep} 
but also enables VLM pretraining~\citep{radford2021learning, align, xu2023metaclip, cherti2023reproducible}.
From the open data sources,
recent methods retrieve task-relevant data to boost performance 
in zero-shot recognition~\citep{liu2023learning, wallingford2023neural, iscen2023retrieval, pmlr-v202-InternetExplorer, parashar2024neglected},
complex reasoning~\citep{pmlr-v119-guu20a, NEURIPS2020_6b493230},
and image generation~\citep{chen2022re, blattmann2022retrieval}.
The recent FSR method~\citep{liu2025few}  retrieves relevant open data to augment the few-shot training data, but it neither evaluates the model's performance on OOD test data nor exploits such data for validation.
We explore methods to leverage such data for validation,
a crucial component that has been underexplored in FSR.
We develop a validation strategy that enables hyperparameter tuning and checkpoint selection, significantly boosting the generalization and robustness of finetuned VLMs for FSR.

{\bf Robustness and Generalization} 
mean different aspects depending on the context,
yet both aim to address real-world scenarios where training and test data are drawn from different distributions~\citep{hendrycks2021many, schmidt2018adversarially}.
For example,
some studies focus on domain adaptation \citep{saito2019semi,berthelot2021adamatch,cao2018partial,you2019universal}, which typically study models' generalization ability by emphasizing adapting a source-domain trained model to a target domain.
Some others study the generalization ability to novel classes by evaluating how a model performs after base-to-novel adaptation~\citep{zhou2022conditional,khattak2023maple,Khattak2024ProText}.
Another line of research investigates the generalization and robustness of models to test data with distribution shift, i.e., those that deviate from the training distribution but still belong to the same classes of training data~\citep{hendrycks2021many, kumar2022finetuning, wortsman2022robust, shi2024lca}.
Similar to this line of research, we study the robustness of FSR methods over ID and OOD test data (drawn from same classes)
and focus on developing realistic validation strategies in the FSR setting.
This demonstrates the stark difference between our work and the current literature of robustness and generalization,
which assumes available validation sets of target-domain or ID data \citep{ren2018learning,sagawa2019distributionally}.

\begin{figure*}[t]
\centering
\small
{\scriptsize
\ 
IN
\hspace{12mm}
IN-V2
\hspace{8mm}
IN-S
\hspace{8mm}
IN-A
\hspace{8mm}
IN-R
\hspace{25mm}
retrieved images}
\hspace{15mm}
\
\\
\vspace{-0.1mm}
  \includegraphics[width = \linewidth, clip=true,trim = 0mm 0mm 0mm 0mm]{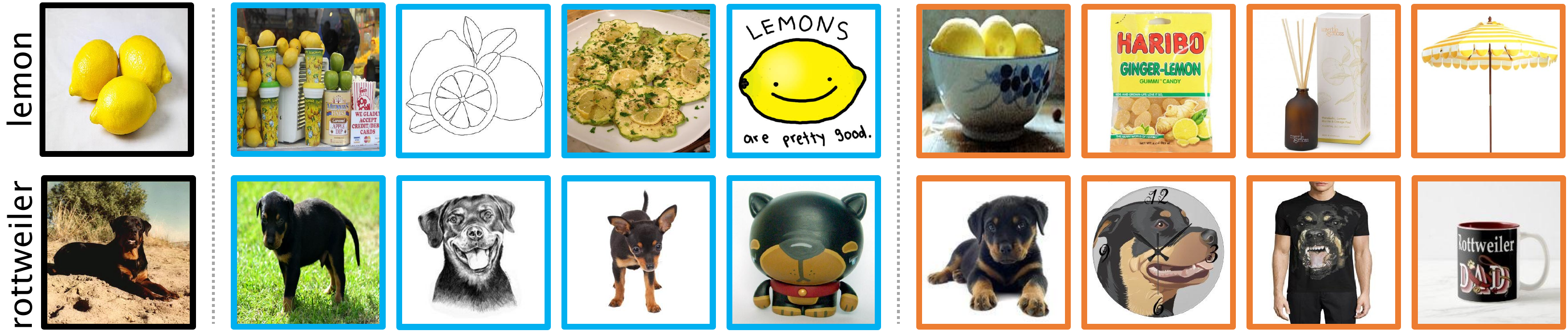}
    \vspace{-6.5mm}
    \caption{\small
    From the ImageNet-OOD benchmark,
    we display two classes with images from 
    the ID training set (marked by black border),
    the OOD test sets (marked by  \textcolor[RGB]{92, 169, 241}{blue border}),
    and the retrieved images from the external data source LAION-400M~\citep{laion400m} (marked by \textcolor[RGB]{219, 133, 50}{orange border}).
    The retrieved examples are not only OOD compared with the ID images, but also noisy due to linguistic ambiguity.
    For example, {\tt lemon} also means a pale yellow color, hence an image of a lemon/yellow umbrella, on the top right, is retrieved.
    }
    \label{fig:imagenet_examples}
    \vspace{-2mm}
\end{figure*}

\section{Validation-Enabled Few-Shot Finetuning}
\label{sec:problem-and-methods}

We begin with the problem formulation of robust FSR (\S\ref{ssec:problem-formulation}).
Then, we introduce a novel validation strategy by studying how to effectively enable checkpoint selection (\S\ref{ssec:checkpoint_selection}),
followed by applying this strategy to layer selection for VLM finetuning (\S\ref{ssec:layer_selection}).
Lastly, we present our final method (\S\ref{ssec:ours}).

\subsection{Problem Formulation}
\label{ssec:problem-formulation}

{\bf Training Protocol.}
In a realistic scenario, 
we have a $K$-way classification task to solve, and an open-source VLM and open data to leverage~\citep{liu2025few}.
The task provides only a small set of labeled data, denoted as  $\mathcal{D} = \{(\I_i, y_i)\}_{i=1}^n$,
where $\I_i$ and $y_i\in \{1, \cdots, K\}$ are the $i^{th}$ image and its label, respectively.
The ``$m$-shot'' setting means that each class has $m$ labeled images; hence $n=m*K$.
The VLM has  a visual encoder $V(\cdot)$ and a text encoder $T(\cdot)$.
We denote the image and text embedding features as $\x_i=V(\I_i)$ and ${\bf t}_i=T(y_i)$, respectively.
FSR methods can adapt the VLM to the task-specific dataset $\mathcal{D}$,
and optionally leverage the open data in training.
Importantly, there is no validation set of task-specific data~\citep{clap24, liu2025few, liu2026solving}.

{\bf Evaluation Protocol.}
We evaluate FSR methods on ID and OOD test sets.
The ID test images follow the same distribution as the task-specific training data, e.g., both are sampled from the same dataset.
The OOD test images have distribution shifts but still capture the same classes as the training data.
Fig.~\ref{fig:imagenet_examples} displays some ID and OOD images.

{\bf Baseline.}
To derive our techniques, we start by finetuning the visual encoder $V(\cdot)$ using the cross-entropy (CE) loss $\mathcal{L}_{CE}$ over the few-shot training data:
$\mathcal{L}_{CE} = \frac{1}{\vert \mathcal{D}\vert } \sum_{(\I_i,y_i) \in \mathcal{D}} 
\ell_{CE}(\I_i, y_i)$, 
where $\ell_{CE}(\I,y) = 
    -\log 
    \frac{\exp(\w_{y}^T  \cdot V(\I))}{\sum_{j=1}^K \exp(\w_{j}^T \cdot V(\I))}$,
and $\w_j$ is the classifier weight for class-$j$ initialized with the corresponding text embedding feature.
Notably, this is a strong baseline, which is reported in  \citep{liu2025few} that outperforms prior PEFT methods.
Quantitative results presented in this section are based on 
the CLIP ViT-B/16 VLM with the ImageNet-based OOD benchmark under the 16-shot setting (refer to \S\ref{sec:exp} for dataset details).

\subsection{Enabling Validation for Checkpoint Selection}
\label{ssec:checkpoint_selection}

\begin{wrapfigure}{r}{0.6\textwidth}
    \centering
    \vspace{-5.5mm}
    \includegraphics[width=\linewidth]{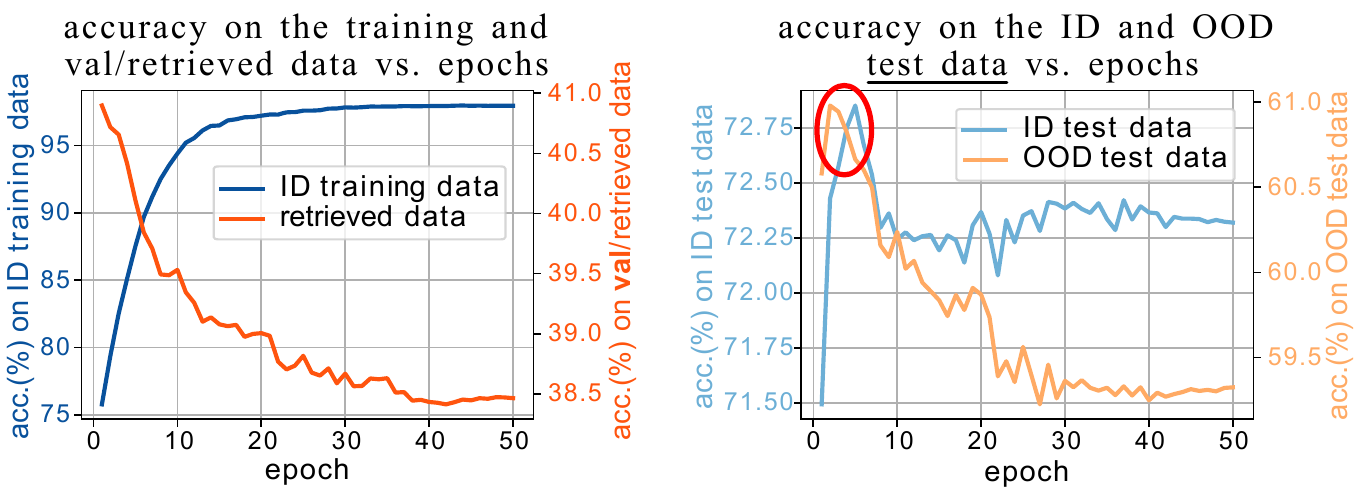} 
    \vspace{-8mm}
    \caption{\small
    {\bf Left: }
    Solely relying on the accuracy on training data or the retrieved data (serving as the validation set) selects the checkpoints at the last epoch or the first epoch.
    {\bf Right: } 
    These checkpoints yield notably lower accuracy on the ID and, especially, the OOD test data than what an oracle would select, e.g., checkpoints at epoch 3-7 marked by red circle.
    Notably, while oracle-selected checkpoints achieve higher accuracy on both ID and OOD test data than most others,
    a single checkpoint rarely achieves the highest accuracy on both simultaneously.
    }
    \label{fig:dilemma}
    \vspace{-2mm}
\end{wrapfigure}

As is well known, validation is crucial for checkpoint selection to avoid overfitting~\citep{prechelt2002early}.
Due to data scarcity in FSR, no validation data practically exist.
Solely relying on the training accuracy (cf. the dark blue curve on Fig.~\ref{fig:dilemma}-left) will select the last checkpoints.
They produce notably degraded ID and OOD test accuracy compared with earlier checkpoints at epochs 3-7 shown in Fig.~\ref{fig:dilemma}-right), which should preferably be selected.
Inspired by \citep{liu2025few}, which retrieves task-relevant external data to augment few-shot training data,
we exploit such data for validation.

\textbf{A Naive Validation Method} is to directly use the retrieved data as the validation set. 
Yet, as the retrieved data is OOD compared with the task-specific training data (Fig.~\ref{fig:imagenet_examples}),
the model finetuned on the ID training data yields degraded performance on this set (Fig.~\ref{fig:dilemma}-left).
As a result,
relying on this validation accuracy selects the original non-finetuned VLM,
preventing effective finetuning toward better generalization.
The analysis from Fig.~\ref{fig:dilemma} reveals that optimal checkpoints 
should strike a balance between two \textit{conflicting objectives}: 
(1) the increase of accuracy on the ID training data, and (2) the decrease of accuracy on the retrieved OOD data.

\begin{figure}[t]
    \begin{minipage}{\textwidth}
        \includegraphics[width=1\linewidth]{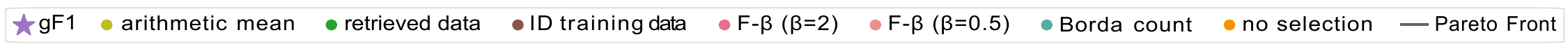} 
    \end{minipage}
    \centering
    \begin{minipage}{0.5\textwidth}
        \includegraphics[width=1\linewidth]{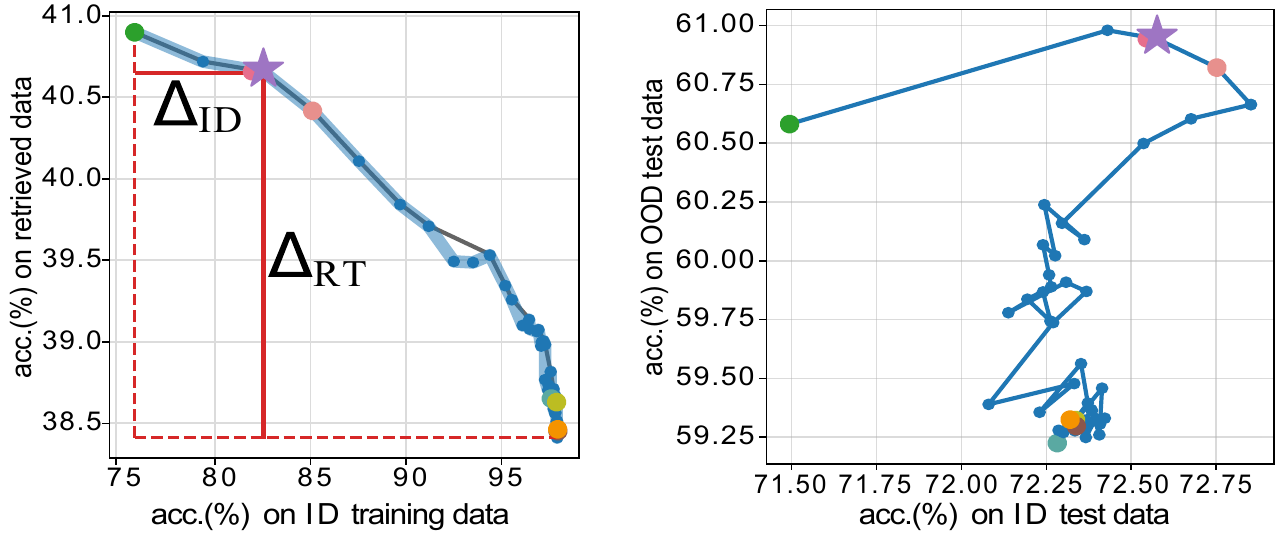} 
    \end{minipage}
    \hfill
    \begin{subfigure}{0.49\textwidth}
    \scalebox{0.6}{
    \setlength{\tabcolsep}{0.55em}
    \begin{tabular}{lccccccc}
    \toprule
    \multirow{2}[2]{*}{\makecell{validation \\ methods}} & \multirow{2}[2]{*}{ckpt} & ID test & \multicolumn{5}{c}{OOD test} \\
    \cmidrule(lr){3-3}
    \cmidrule(lr){4-8}
    &  & \cellcolor{col11}IN & \cellcolor{col33}avg. & IN-V2 & IN-S & IN-A & IN-R \\
    \midrule
    \NoSelection no selection & 50 & \cellcolor{col11}72.32 & \cellcolor{col33}59.33 & 64.85 & 48.10 & 49.03 & 75.33 \\
    \IDtrn ID train data & 44 & \cellcolor{col11}72.34 & \cellcolor{col33}59.30 & 64.83 & 48.10 & 48.95 & 75.32 \\
    \Rtr retrieved data & \ \ 1 & \cellcolor{col11}71.49 & \cellcolor{col33}60.58 & 64.49 & \underline{49.10} & 50.31 & \textbf{78.42} \\
    \ArithmeticMean arithmetic mean & 35 & \cellcolor{col11}72.34 & \cellcolor{col33}59.32 & 64.82 & 48.11 & 49.05 & 75.31 \\
    \BordaCount Borda count  & 27 & \cellcolor{col11}72.28 & \cellcolor{col33}59.23 & 64.66 & 47.97 & 48.79 & 75.48 \\
    \FHALF F-$\beta$ ($\beta=0.5$) & \ \ 4 & \cellcolor{col11}\textbf{72.75} & \cellcolor{col33}\underline{60.82} & \underline{65.53} & 49.02 & \textbf{51.04} & 77.69 \\
    \FTWO F-$\beta$ ($\beta=2$) & \ \ 3 & \cellcolor{col11}\underline{72.58} & \cellcolor{col33}\textbf{60.94} & \textbf{65.54} & \textbf{49.39} & 50.72 & \underline{78.12} \\
    \gFONE gF1 & \ \ 3 & \cellcolor{col11}\underline{72.58} & \cellcolor{col33}\textbf{60.94} & \textbf{65.54} & \textbf{49.39} & 50.72 & \underline{78.12} \\

    \bottomrule
    \end{tabular}}
    \end{subfigure}
    \vspace{-1mm}
    \caption{\small
    {\bf Left:} Maximizing the gF1 score selects the epoch-3 checkpoint. 
    {\bf Right:} We compare this checkpoint with others selected by different methods:
    the arithmetic mean of accuracies on the ID training data and retrieved data,
    the accuracy on the retrieved data only,
    the accuracy on the ID training data only,
    the rank aggregation method Borda count~\cite{aslam2001models}, 
    F-$\beta$ score,
    and no selection (i.e., using the last checkpoint).
    Clearly,  gF1's selection performs robustly well and generalizes to both ID and OOD test data. 
    }
    \label{fig:validation}
    \vspace{-1mm}
\end{figure}

{\bf Multi-Objective Optimization (MOO)} aims to maximize or minimize two or more conflicting objective functions at the same time
\citep{sener2018multi, zafar2017fairness}.
It precisely describes the checkpoint selection problem that should jointly consider the above two conflicting objectives.
MOO produces a set of optimal trade-offs known as the Pareto optimal set and its graphical representation, the Pareto front.
In our problem, the Pareto front is depicted in Fig.~\ref{fig:validation}-left, with the accuracy of the training and retrieved/validation data as two axes / objectives.
Thus, designing a validation method amounts to designing a mechanism to select an operating point on the Pareto front.

{\bf Operating Point Selection} on a Pareto front is fundamentally a policy decision \citep{ngatchou2005pareto, sener2018multi, zafar2017fairness}.
Without accounting for the Pareto front,
a naive baseline is to take the arithmetic mean of the accuracies on the training data and retrieved validation data.
However, the arithmetic mean cannot capture points in non-convex regions of a Pareto front \citep{das1997closer, vandenberghe2004convex} and implicitly assumes equal weighting between the two accuracies.
Consequently,
when training accuracy dominates validation accuracy (Fig.~\ref{fig:dilemma}-left), the arithmetic mean favors an overfitting checkpoint late in  training.
To address these limitations,
we propose using the harmonic mean (F1 score),
which is guaranteed to select a Pareto-optimal point, as described by the following theorem (ref. Appendix \S\ref{sec:proof} for a  formal proof):
\begin{theorem}\label{theorem1}
Let $\mathcal{F} \subset \mathbb{R}_{>0}^n$ be a feasible set of strictly positive objective vectors. If $\mathbf{x}^* \in \mathcal{F}$ is a global maximizer of the harmonic mean over $\mathcal{F}$, i.e.,
$H(\mathbf{x}^*) \ge H(\mathbf{x}) \quad \forall \mathbf{x} \in \mathcal{F}$,
then $\mathbf{x}^*$ is Pareto-optimal.
\end{theorem}

Other than Pareto optimality,
the F1 score has other desirable properties.
For example, when the gap between the two accuracies gets bigger,
the F1 score gets increasingly governed by the weaker metric,
meaning further gains on the stronger metric yield diminishing returns.
Moreover, we subtract the min-accuracy,
thought of as reference baselines,
from each metric to align the contour focal point with the actual 
trade-off boundary (Fig.~\ref{fig:validation}-left and
Table~\ref{tab:min-subtraction}) \citep{miettinen1999nonlinear, marler2004survey}. 
This shift in reference baselines does not alter the
\begin{wrapfigure}{r}{0.45\textwidth}
{
\setlength{\tabcolsep}{0.3em} 
\centering
\vspace{-3mm}
\captionof{table}{\small
Min-subtraction aligns the contour focal point with the actual trade-off boundary (Fig.~\ref{fig:validation}-left),
leading to more robust validation.
}
\label{tab:min-subtraction} 
\vspace{-3mm}
\scalebox{0.75}{
\begin{tabular}{lcccccc}
\toprule
\multirow{2}[2]{*}{\makecell{harmonic \\ mean}} 
& ID test & \multicolumn{5}{c}{OOD test} \\
\cmidrule(lr){2-2} \cmidrule(lr){3-7}
&  \cellcolor{col11}IN & \cellcolor{col33}avg. & IN-V2 & IN-S & IN-A & IN-R \\
\midrule
w/o min-sub & \cellcolor{col11}72.24 & \cellcolor{col33}60.24 & 65.21 & 48.36 & \textbf{50.95} & 76.43 \\
w/ min-sub & \cellcolor{col11}\textbf{72.58} & \cellcolor{col33}\textbf{60.94} & \textbf{65.54} & \textbf{49.39} & 50.72 & \textbf{78.12} \\
\bottomrule
\end{tabular}}
\vspace{-2mm}
}
\end{wrapfigure}
underlying Pareto front but restores
the harmonic mean's ability to penalize imbalanced trade-offs when
metrics have high baseline values (e.g., the training accuracy) \citep{flach2015precision, powers2020evaluation}.
We validate the robustness of gF1 to perturbations of the reference baselines later in Fig.~\ref{fig:layer_selection_sensitivity}. 
In plain language,
we compute the F1 score over (1) the performance gain on the ID training data,
and (2) the performance preservation on the retrieved data (Fig.~\ref{fig:validation}-left).

{\bf The proposed gF1 score.}
Following the above,
we name the metric for validation {\bf gF1} score,
standing for a measure of model generalization based on the F1 score.
Concretely,
for the $t^{th}$ checkpoint, we record its training accuracy ($\text{acc}^t_{ID}$) and the accuracy ($\text{acc}^t_{RT}$) on the retrieved data which serve as the validation set.
We place all the checkpoints on a 2D plane with $x$ and $y$ specifying the two accuracies (Fig.~\ref{fig:validation}-left).
Then,
for the $t^{th}$ checkpoint, 
we compute its ``performance gain'' on the ID training data as $\Delta_{ID}^t= \text{acc}^t_{ID} - \min_i(\text{acc}^i_{ID})$,
and  ``performance preservation'' on the retrieved data as $\Delta_{RT}^t=\text{acc}^t_{RT}-\min_i(\text{acc}^i_{RT})$.
The gF1 score is computed as 
\begin{equation}\small
   \text{gF1}^t = 2 \times   \frac{\Delta_{ID}^t \times \Delta_{RT}^t}{\Delta_{ID}^t + \Delta_{RT}^t}
   \label{eq:gF1}
\end{equation}
Fig.~\ref{fig:validation}-left shows that the gF1 score selects the checkpoint at epoch-3.
Fig.~\ref{fig:validation}-right verifies 
that the selected checkpoint outperforms that selected by alternative methods and generalizes robustly to OOD test data with various distribution shifts.

\textbf{Remarks.}
Our gF1 could be extended with given prior knowledge.
For example, 
if domain similarity between retrieved data and task-specific ID data could be measured,
one would assign different weights to the two objectives in Eq.~\ref{eq:gF1} and thus use the F-$\beta$ metric.
One may think about tuning $\beta$, but this requires a separate validation set, 
which is inaccessible in the realistic few-shot setup.
Nevertheless,
Fig.~\ref{fig:validation} verifies that gF1 generalizes more robustly than alternative measures for checkpoint selection.
Lastly, gF1 introduces no extra computational cost over the conventional validation pipeline, which uses validation accuracy:
gF1 is simply a replacement of this validation accuracy.

\subsection{Enabling Validation for Layer Selection}
\label{ssec:layer_selection}

The gF1 score enables tuning various hyperparameters;
we use it to tune an underexplored one, determining what parameters in a pretrained model to finetune.
Intuitively, finetuning too few parameters insufficiently adapts the pretrained model to the downstream task, 
while finetuning too many easily overfits the few-shot training data, distorting pretrained features and degrading generalization~\citep{kumar2022finetuning}.
Moreover, low-level layers of pretrained parameters are more generic,
while high-level ones are more tailored to downstream tasks.
Hence,
we extend our validation strategy to determine $k$ for {\bf partially finetuning (PFT)} the top $k$ layers of the pretrained model $V(\cdot)$.

\begin{figure}[t]
  \begin{minipage}{0.49\textwidth}
  \centering
  \vspace{0pt}
    \includegraphics[width=\linewidth]{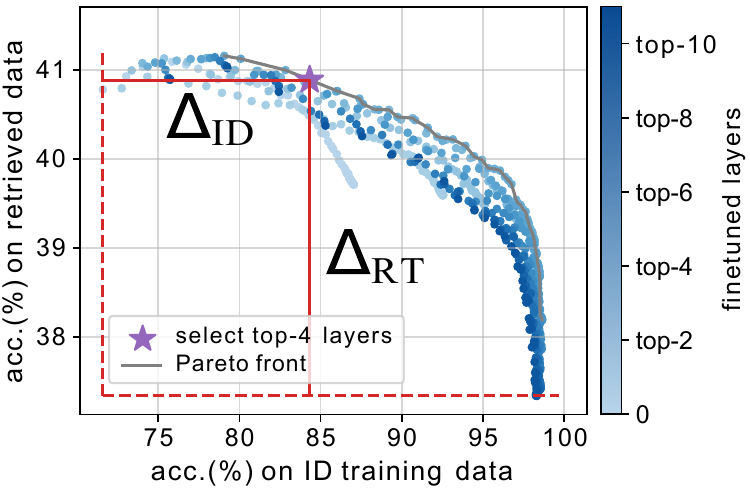}
    \end{minipage}
  \hfill
  \begin{minipage}{0.49\textwidth}
      \centering
    \scalebox{0.65}{
        \begin{tabular}{lcccccc}
            \toprule
            \multicolumn{1}{l}{\multirow{2}[4]{*}{finetuned layers}} & ID test   & \multicolumn{5}{c}{OOD test} \\
        \cmidrule(lr){2-2}
        \cmidrule(lr){3-7}
        & \cellcolor{col11}IN    & \cellcolor{col33}avg.  & IN-V2 & IN-S  & IN-A  & IN-R \\
            \midrule
            \rowcolor{lightlightgrey}linear probing & \cellcolor{col11}71.52  & \cellcolor{col33}60.65  & 64.48  & 49.10  & 50.79  & 78.21  \\
            \rowcolor{lightlightgrey}FFT & \cellcolor{col11}72.32 & \cellcolor{col33}59.33 & 64.85 & 48.10 & 49.03 & 75.33 \\
            
            \midrule
            
            top-1 & \cellcolor{col11}72.30  & \cellcolor{col33}61.17  & 65.44  & 49.36  & 51.29  & 78.57  \\
            top-2 & \cellcolor{col11}72.89  & \cellcolor{col33}61.70  & 66.07  & 49.81  & 51.91  & 79.01  \\
            top-3 & \cellcolor{col11}73.19  & \cellcolor{col33}\underline{61.78}  & 66.15  & \textbf{49.95}  & \underline{51.96}  & \textbf{79.06}  \\
            top-4$^*$ & \cellcolor{col11}\underline{73.35}  & \cellcolor{col33}\textbf{61.80}  & \underline{66.23}  & \underline{49.89}  & \textbf{52.04}  & \underline{79.02}  \\
            top-5 & \cellcolor{col11}\textbf{73.41}  & \cellcolor{col33}61.74  & 66.46  & 49.79  & 51.95  & 78.78  \\
            top-6 & \cellcolor{col11}73.40 & \cellcolor{col33}61.61 & \textbf{66.49} & 49.60 & 51.79 & 78.55 \\
            top-7 & \cellcolor{col11}73.20  & \cellcolor{col33}61.28  & 66.22  & 49.35  & 51.41  & 78.14  \\
            top-8 & \cellcolor{col11}73.03 & \cellcolor{col33}61.02 & 65.98 & 49.21 & 51.16 & 77.74 \\
            top-9 & \cellcolor{col11}72.99  & \cellcolor{col33}60.90  & 65.95  & 49.14  & 50.91  & 77.58  \\
            top-10 & \cellcolor{col11}72.89 & \cellcolor{col33}60.91 & 65.90 & 49.11 & 50.97 & 77.68 \\
            top-11 & \cellcolor{col11}72.83 & \cellcolor{col33}60.92 & 65.87 & 49.02 & 51.19 & 77.59 \\
            \bottomrule
            \end{tabular}}%
        \label{tab:layer_selection}%
  \end{minipage}
  \vspace{-1mm}
  \caption{\small
  {\bf Layer selection via the proposed gF1.} 
  {\bf Left:}
  For each $k$, 
  we partially finetune (PFT) the top-$k$ layers of the pretrained model $V(\cdot)$ and collect the intermediate checkpoints. 
  Each checkpoint is mapped to a 2D coordinate defined by its accuracy on the ID training set and retrieved data.
  Across all checkpoints and values of $k$, we select the point maximizing the gF1 score (Eq.~\ref{eq:gF1}).
  This selected point, $k=4$, is guaranteed to lie on the Pareto front, highlighted by gray curve.
  {\bf Right:}
  To validate gF1-based layer selection,
  we report  the ID and OOD test accuracies of the checkpoints independently selected by gF1 for each $k$.
  The results confirm that finetuning the top-4 layers achieves substantial accuracy gains over  linear probing and full finetuning.
  }
    \vspace{-1mm}
    \label{fig:layer-selection}
\end{figure}

Concretely,
for a given $k$, we PFT $V(\cdot)$ and save intermediate checkpoints (e.g., at the end of each epoch).
Each checkpoint is mapped to a 2D coordinate defined by its accuracy on the ID training data and the retrieved data.
Plotting all checkpoints across all values of $k$ yields the scatter plot in Fig.~\ref{fig:layer-selection}-left.
From this set, we select the checkpoint that maximizes the gF1 score, which simultaneously identifies the layer depth for PFT ($k=4$ in this example).
The selected checkpoint lies on the Pareto front (ref. Theorem \ref{theorem1}).
Fig.~\ref{fig:layer-selection}-right empirically 
validates the gF1-based layer selection
by comparing the ID and OOD test accuracies of independently gF1-selected checkpoint for each $k$.
Appendix \S\ref{sec:more_gF1_results} provides results on tuning other hyperparameters using gF1.

\begin{figure}[h]
    \begin{minipage}{.48\textwidth}
        \centering
        \vspace{-4.2mm}
        \includegraphics[width=0.99\linewidth]{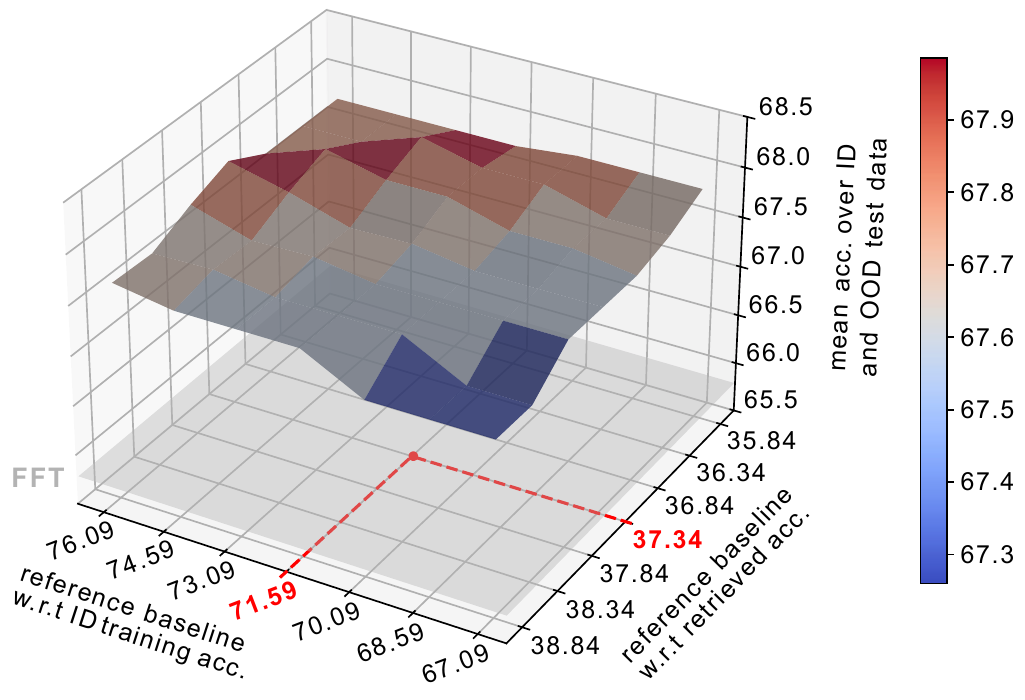} 
        \vspace{-2mm}
    \end{minipage}
    \hfill
    \begin{minipage}{.51\textwidth}
        \caption{\small
    {\bf Robustness of gF1 to perturbed reference baselines.} 
    Based on Fig.~\ref{fig:layer-selection}-left,
    we perturb the reference baselines, i.e., assuming different min-accuracies used in gF1 calculation (Eq.~\ref{eq:gF1}).
    We report these gF1-selected checkpoints w.r.t the mean accuracy on ID and OOD test data.
    The original min-accuracies / reference baselines are marked in red.
    The results, compared with the strong method FFT \citep{liu2025few} (gray surface),
    validate the robustness of gF1 to a broad range of reference baselines for selecting Pareto-optimal points that yield stable performance in test time.
    }
    \label{fig:layer_selection_sensitivity}
    \end{minipage}
    \vspace{-7mm}
\end{figure}

{\bf Remarks.} Fig.~\ref{fig:layer-selection} demonstrates joint layer and checkpoint selection by comparing all candidates via the gF1 metric to identify a single Pareto-optimal point.
However, this joint paradigm requires
all intermediate checkpoints to be retained, incurring a significant storage overhead.
To mitigate this,
we adopt a two-step paradigm:
(1) perform independent checkpoint selection for each layer depth $k$ using gF1 (\S\ref{ssec:checkpoint_selection}),
and 
(2) re-evaluate the gF1 scores of these selected checkpoints under a unified reference baseline  (i.e., shared minimum accuracies) and return the checkpoint corresponding to the highest gF1.
This two-step paradigm retains only a single checkpoint per layer configuration, drastically reducing storage requirements.
\Cref{theorem2} and \Cref{corollary1} show that this two-step procedure produces a globally Pareto-optimal checkpoint.
It is worth noting that, in Step 2, directly comparing gF1 scores across different $k$ right after Step 1 is invalid,
because each layer depth traces a distinct optimization trajectory with a unique reference baseline;
hence, Step 2 is essential.
Furthermore,
while the checkpoint yielded by the two-step paradigm remains Pareto-optimal,
it may differ from the joint-selected checkpoint due to the shifted reference baselines.
Nevertheless,
these different checkpoints via  gF1  consistently achieve robust performance,
as validated in Fig.~\ref{fig:layer_selection_sensitivity}.

\begin{theorem}
\label{theorem2}
Let $\mathcal{F} \subset \mathbb{R}_{>0}^n$ be a feasible set of strictly positive objective vectors.
Let $\{\mathcal{F}_i\}_{i \in \mathcal{I}}$ be an arbitrary collection of subsets of $\mathcal{F}$ indexed by a set $\mathcal{I}$,
and let $\mathcal{F}  = \bigcup_{i \in \mathcal{I}} \mathcal{F}_i$ be their union.
Then the Pareto front of $\mathcal{F}$ is a subset of the union of the Pareto front of $\mathcal{F}_i$: $
\mathcal{P}\left(\bigcup_{i \in \mathcal{I}} \mathcal{F}_i\right) \subseteq \bigcup_{i \in \mathcal{I}} \mathcal{P}(\mathcal{F}_i)$.
\end{theorem}
\vspace{-2mm}

\begin{corollary}
\label{corollary1}
Suppose that $\mathcal{I}$ is non-empty finite index set
and each subset $\mathcal{F}_i$ ($i\in\mathcal{I}$) is non-empty and finite (or compact). 
Then 
the global Pareto front $\mathcal{P}\left(\bigcup_{i \in \mathcal{I}} \mathcal{F}_i\right)$ can be recovered directly from the union of local Pareto fronts: 
$\mathcal{P}\left(\bigcup_{i \in \mathcal{I}} \mathcal{F}_i\right) = \mathcal{P}\left(\bigcup_{i \in \mathcal{I}} \mathcal{P}(\mathcal{F}_i)\right)$.
\end{corollary}

\subsection{VEST: Validation-Enabled Stage-wise Tuning}
\label{ssec:ours}

With the gF1-based validation strategy,
we present the final method that retrieves external data for both training data augmentation and validation within a stage-wise finetuning paradigm.
We call this method Validation-Enabled Stage-wise Tuning (VEST).

{\bf Retrieval-based Augmentation (RA)} is a modern philosophy that leverages open data source
to solve downstream tasks~\citep{liu2023learning, wallingford2023neural, parashar2024neglected, liu2025few, liu2026solving}.
It retrieves examples relevant to the $K$ classes concerned by the downstream task, and uses them to augment training data.
Importantly, we further exploit the retrieved data with our validation strategy to determine what parameters to finetune and 
select checkpoints.
Let's denote the set of retrieved data as $\mathcal{D}_R = \{(\I_i, y_i)\}_{i=1}^N$.
We compute the CE loss over it:
$\mathcal{L}_{RA} = \frac{1}{\vert \mathcal{D}_R\vert } \sum_{(\I_i, y_i)\in\mathcal{D}_R}\ell_{CE}(\I_i,y_i)$.

{\bf One-stage training.}
With RA, one can jointly minimize the losses to partially finetune the visual encoder on both few-shot and retrieved data:
$\mathcal{L} = \mathcal{L}_{CE} + \mathcal{L}_{RA}.$
Notably, the retrieved examples have label noises, distributional shifts compared with task-specific training images (Fig.~\ref{fig:imagenet_examples}),
and imbalanced class distributions~\citep{parashar2024neglected, liu2025few}.
Consequently, 
this one-stage finetuning provides limited gains w.r.t either ID or OOD test accuracy  (\Cref{tab:components}),
motivating our stage-wise finetuning pipeline, presented next.

{\bf Stage-wise training} is shown to mitigate issues caused by data imbalance~\citep{kang2019decoupling} and domain gaps~\citep{liu2025few}, and serves as a transfer learning paradigm~\citep{mahajan2018exploring}.
Motivated by these benefits, VEST adopts a three-stage pipeline, with a layer selection stage preceding two finetuning stages. In Stage 0, the proposed validation method selects the PFT layers. In Stage 1, the selected layers are finetuned on both few-shot ID and retrieved data. In Stage 2, they are further finetuned on few-shot ID data. Compared with one-stage training, VEST achieves higher ID and OOD accuracy while substantially reducing training cost (\Cref{tab:components}).

\begin{figure*}[t]
  \centering
  \small
  \vspace{-1mm}
  \includegraphics[width=1.0\linewidth, clip=true, trim = 0mm 0mm 0mm 0mm]{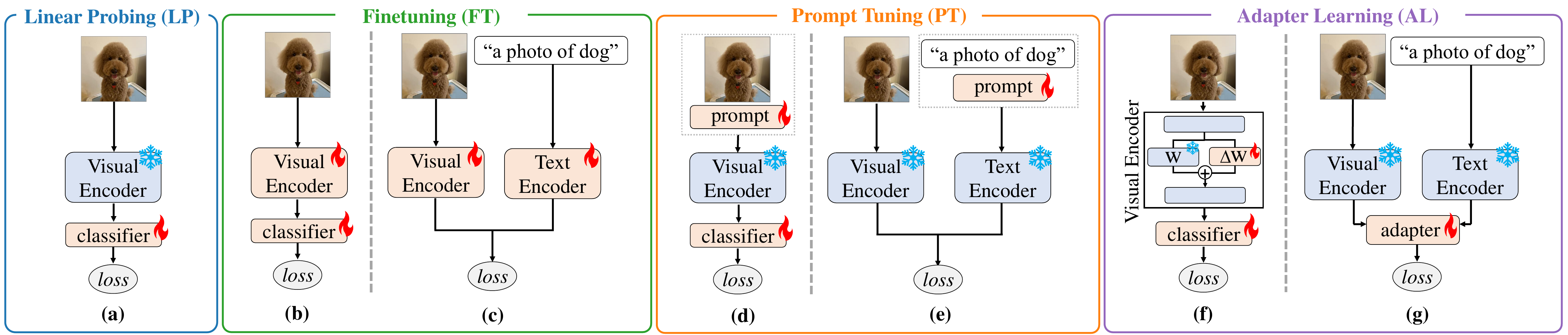}
  \vspace{-4mm}
  \caption{\small
  {\bf Diagrams of prevailing VLM adaptation methods.}
    We categorize them into four groups.
    \textbf{(a)}
    \textcolor[RGB]{31, 119, 180}{Linear Probing (LP)} learns a linear classifier on top of a pretrained visual encoder~\citep{radford2021learning}.
    \textcolor[RGB]{44, 160, 44}{Finetuning (FT)} learns to update either \textbf{(b)}  the visual encoder via a cross entropy loss~\citep{liu2025few} or \textbf{(c)} the entire VLM via a contrastive loss~\citep{goyal2023finetune}.
    \textcolor[RGB]{255, 127, 14}{Prompt Tuning (PT)}  learns either \textbf{(d)} visual prompts~\citep{jia2022visual} or \textbf{(e)} text prompts~\citep{zhou2022learning} in the input space with a frozen VLM.
    \textcolor[RGB]{148, 103, 189}{Adapter Learning (AL)}  freezes the VLM and learns lightweight modules, such as \textbf{(f)} low-rank matrices~\citep{hu2022lora, zanella2024low}  or \textbf{(g)} some layers of parameters~\citep{zhang2022tip}.
    }
  \vspace{-1mm}
  \label{fig:vlm_adapt}
\end{figure*}

\section{Experiments}
\label{sec:exp}

Through extensive experiments, we validate the effectiveness of our validation strategy, analyze the stages in our VEST, and demonstrate its state-of-the-art performance.
We first describe the datasets and VLMs in experiments, introduce the compared methods, and present important details of implementation.
We then carry out in-depth analyses of our method and benchmark more approaches.

\subsection{Experimental Setups}

\textbf{Benchmarks} include the ImageNet-based OOD benchmark,
DomainNet  \citep{peng2019moment}, 
Living17 \citep{santurkar2020breeds}, Entity30 and fMoW  \citep{christie2018functional}.
Due to limited space and the fact that conclusions hold across benchmarks,
we report results on the largest-scale ImageNet-based one in the main paper
and provide results on the rest in Appendix \S\ref{sec:other_datasets}.
This benchmark uses ImageNet (IN) \citep{deng2009imagenet} to randomly sample few-shot training data of 1,000 classes as the ID training data,
and uses IN's official validation set as the ID test set.
It constructs OOD test sets based on IN's OOD variants, 
including ImageNet-V2 (IN-V2)~\citep{recht2019imagenet}, 
ImageNet-Sketch (IN-S)~\citep{wang2019learning}, 
ImageNet-Adversarial (IN-A)~\citep{hendrycks2021natural}, 
and ImageNet-Rendition (IN-R)~\citep{hendrycks2021many}.
IN is released for non-commercial research and educational purposes,
and its OOD variants are open-source under the MIT License.
Refer to Fig.~\ref{fig:imagenet_examples} for some visual examples,
and Appendix \S\ref{sec:datasets} for more details of these datasets.

\textbf{Models and Protocols.}
Following the literature~\citep{lin2023multimodality, clap24}, 
we use CLIP ViT-B/16 as the pretrained VLM, whose pretraining data have no overlap with the above ImageNet-related datasets~\citep{radford2021learning}.
Following \citep{liu2025few}, we use LAION-400M~\citep{laion400m} as the external data resource and their method to retrieve task-relevant data.
LAION-400M is open-source under CC-BY 4.0 License and has weak affinity to ImageNet~\citep{cherti2023reproducible}.
For each class of ImageNet, we retrieve $\sim$500 images to augment the few-shot labeled data,
and another $\sim$50 images to construct our validation set.
Retrieval details for other benchmarks are provided in Appendix \S\ref{sec:retrieval}.
We conduct experiments under  4-, 8-, and 16-shot settings, and the ablation study in the 16-shot setting.

{
\setlength{\tabcolsep}{0.45em}
\begin{table}[t]
\small
  \centering
  \caption{\small 
    {\bf Benchmarking results} w.r.t top-1 accuracy on
    the ID test set ImageNet (IN) and the OOD test sets ImageNet-V2 (IN-V2), ImageNet-Sketch (IN-S), ImageNet-A (IN-A), and ImageNet-Rendition (IN-R).
    Zero-shot methods serve as references.
    For convenient comparison,  for each method, we report averaged accuracy (marked in {\setlength{\fboxsep}{1pt}\colorbox{col33}{blue}}) across the four OOD test sets.
    We run all the methods with their original validation protocols over the CLIP ViT-B/16 model~\citep{radford2021learning}, under 16-, 8-, and 4-shot settings.
    Numbers for marked methods are copied from their papers.
    Particularly, for methods that originally exploit a validation set of ID data for tuning, we use the accuracy on the ID test set to tune them, reporting their over-optimistic results and marking them as ``{\color{gray}ID test acc.}''.
    Both SWAT and VEST adopt RA, using the same retrieved data during training.
    Enabled by our validation strategy,
    VEST resoundingly outperforms SWAT  on the ID and OOD test data.
    }
      \label{tab:baseline}  
  \vspace{-3mm}
  \scalebox{0.76}{
    \begin{tabular}{llllcccccccccc}
    \toprule
          &  &  &  & \multicolumn{6}{c}{16-shot} & \multicolumn{2}{c}{8-shot} & \multicolumn{2}{c}{4-shot} \\
    \cmidrule(lr){5-10}
    \cmidrule(lr){11-12}
    \cmidrule(lr){13-14}
    & & & & ID  & \multicolumn{5}{c}{OOD} & ID & OOD & ID  & OOD  \\
    \cmidrule(lr){5-5}
    \cmidrule(lr){6-10}
    \cmidrule(lr){11-11}
    \cmidrule(lr){12-12}
    \cmidrule(lr){13-13}
    \cmidrule(lr){14-14}
    &  method & venue \& year & validation & \cellcolor{col11}IN & \cellcolor{col33}avg. & IN-V2 & IN-S  & IN-A  & IN-R & \cellcolor{col11}IN & \cellcolor{col33}avg. & \cellcolor{col11}IN & \cellcolor{col33}avg.\\
    \midrule
     \multirow{2}{*}{\makecell{Zero-Shot}} & Prompting & {ICML'21} & -- & \cellcolor{col11}66.75 & \cellcolor{col33}57.15 & 60.91 & 46.16 & 47.53 & 73.99  & \cellcolor{col11}66.75 & \cellcolor{col33}57.15 & \cellcolor{col11}66.75 & \cellcolor{col33}57.15 \\

     & CuPL & {ICCV'23}  & -- & \cellcolor{col11}69.64 & \cellcolor{col33}60.04 & 63.37 & 49.03 & 50.68 & 77.07 & \cellcolor{col11}69.64 & \cellcolor{col33}60.04 & \cellcolor{col11}69.64 & \cellcolor{col33}60.04 \\
    
    \midrule
       
    \multirow{7}{*}{\makecell{Prompt \\ Tuning}} & CoOp & {IJCV'22}  & {\color{gray}ID test acc.} & \cellcolor{col11}71.69 & \cellcolor{col33}59.24 & 64.34 & 47.20  & 50.20  & 75.21 & \cellcolor{col11}70.80 & \cellcolor{col33}59.68 & \cellcolor{col11}70.29 & \cellcolor{col33}59.74 \\
    & CoCoOp$^*$ & {CVPR'22} & {\color{gray}ID test acc.} & \cellcolor{col11}71.02 & \cellcolor{col33}59.91 & 64.07 & 48.75 & 50.63 & 76.18 & \cellcolor{col11}-- & \cellcolor{col33}-- & \cellcolor{col11}-- & \cellcolor{col33}--\\
    & MaPLe & {CVPR'23} & {\color{gray}ID test acc.} & \cellcolor{col11}71.15 & \cellcolor{col33}60.42 & 64.49 & 48.69 & 50.91 & 77.58 & \cellcolor{col11}70.76 & \cellcolor{col33}60.23 & \cellcolor{col11}70.58 & \cellcolor{col33}60.13 \\
    & VPT{ shallow} & {ECCV'22} & {\color{gray}ID test acc.} & \cellcolor{col11}72.75 & \cellcolor{col33}60.80 & 65.70 & 49.33 & 49.68 & 78.56 & \cellcolor{col11}71.48 & \cellcolor{col33}60.29 & \cellcolor{col11}70.31 & \cellcolor{col33}59.76 \\
    & VPT{ deep} & {ECCV'22} & {\color{gray}ID test acc.} & \cellcolor{col11}72.78 & \cellcolor{col33}59.12 & 65.04 & 47.83 & 47.11 & 76.49 & \cellcolor{col11}71.22 & \cellcolor{col33}59.30 & \cellcolor{col11}70.54 & \cellcolor{col33}59.71 \\ 
  & ProText$^*$ & {AAAI'25} & {\color{gray}ID test acc.} & \cellcolor{col11}70.22 & \cellcolor{col33}60.45 & 63.54 & 49.45 & 51.47 & 77.35 & \cellcolor{col11}-- & \cellcolor{col33}-- & \cellcolor{col11}-- & \cellcolor{col33}-- \\
  & Promise$^*$ & {ICLR'26} & {\color{gray}ID test acc.} & \cellcolor{col11}72.77 & \cellcolor{col33}62.23 & 66.92 & 50.28 & \underline{53.21} & 78.52 & \cellcolor{col11}-- & \cellcolor{col33}-- & \cellcolor{col11}-- & \cellcolor{col33}-- \\

    \midrule

    \multirow{6}{*}{\makecell{Adapter \\ Learning}} & Tip-Adapter & {ECCV'22} & {\color{gray}ID test acc.} & \cellcolor{col11}70.51 & \cellcolor{col33}60.14 & 63.23 & 48.64 & 50.88 & 77.80 & \cellcolor{col11}70.07 & \cellcolor{col33}60.05 & \cellcolor{col11}69.70 & \cellcolor{col33}60.05 \\
    & Tip-Adapter-F & {ECCV'22} & {\color{gray}ID test acc.} & \cellcolor{col11}73.40 & \cellcolor{col33}59.85 & 65.43 & 47.95 & 49.08 & 76.95 & \cellcolor{col11}71.77 & \cellcolor{col33}59.91 & \cellcolor{col11}70.74 & \cellcolor{col33}59.84 \\
    & CLIP-Adapter & {IJCV'23} & {\color{gray}ID test acc.} & \cellcolor{col11}71.61 & \cellcolor{col33}58.03 & 64.05 & 46.66 & 47.87 & 73.53 & \cellcolor{col11}70.25 & \cellcolor{col33}57.74 & \cellcolor{col11}69.55 & \cellcolor{col33}58.36 \\
    & TaskRes & {CVPR'23} & {\color{gray}ID test acc.} & \cellcolor{col11}73.48 & \cellcolor{col33}59.95 & 65.10  & 48.12 & 49.96 & 76.62 & \cellcolor{col11}71.91 & \cellcolor{col33}60.04 & \cellcolor{col11}71.28 & \cellcolor{col33}59.51 \\
          
    & DMN-ZS$^*$ & {CVPR'24} & {\color{gray}ID test acc.}  & \cellcolor{col11}72.25 & \cellcolor{col33}\underline{63.80} & 65.17 & 53.20 &  \textbf{58.28} & 78.55
    & \cellcolor{col11}--   &  \cellcolor{col33}--   &  \cellcolor{col11}--   &  \cellcolor{col33}-- 
    \\

    & LoRA & {ICLR'22}  & {\color{gray}ID test acc.}
    & \cellcolor{col11}72.89 & \cellcolor{col33}60.63 & 65.31 & 48.40 & 50.67 & 78.12 & \cellcolor{col11}71.82 & \cellcolor{col33}60.36 & \cellcolor{col11}70.62 & \cellcolor{col33}59.88 \\

    \midrule

   \multirow{5}{*}{\makecell{Linear \\ Probing}} & LP & {ICML'21}    & {\color{gray}ID test acc.} & \cellcolor{col11}73.36 & \cellcolor{col33}60.53  & 65.33 &  48.91 & 49.76 & 78.12 & \cellcolor{col11}72.45 & \cellcolor{col33}59.77 & \cellcolor{col11}71.44 & \cellcolor{col33}59.55 \\
    & LP w/ WiSE-FT & {CVPR'22} & {\color{gray}ID test acc.} & \cellcolor{col11}71.97 & \cellcolor{col33}60.73 & 64.47 & 49.43 & 50.69 & 78.33 & \cellcolor{col11}71.37 & \cellcolor{col33}60.48 & \cellcolor{col11}70.82 & \cellcolor{col33}60.20 \\

    & CMLP & {CVPR'23} & {\color{gray}ID test acc.} & \cellcolor{col11}73.32  & \cellcolor{col33}60.11 & 64.88 & 48.41 & 49.58 & 77.56 & \cellcolor{col11}72.21 & \cellcolor{col33}59.93 & \cellcolor{col11}70.90 & \cellcolor{col33}59.56 \\
    & LP w/ LCA & {ICML'24} & {\color{gray}ID test acc.} & \cellcolor{col11}69.30  & \cellcolor{col33}59.60 & 62.41 & 48.52 & 49.65 & 77.81 & \cellcolor{col11}69.23 & \cellcolor{col33}59.48 & \cellcolor{col11}69.15 & \cellcolor{col33}59.61 \\

    & CLAP & {CVPR'24}  & {\color{black}val-free} & \cellcolor{col11}72.97 & \cellcolor{col33}59.86 & 65.11 & 48.00 & 49.53 & 76.81 & \cellcolor{col11}71.56 & \cellcolor{col33}59.53 & \cellcolor{col11}70.91 & \cellcolor{col33}59.17 \\

    \midrule

    \multirow{6}{*}{Finetuning}       
    & FLYP & {CVPR'23}  & {\color{gray}ID test acc.} &  \cellcolor{col11}\underline{74.33} & \cellcolor{col33}58.24 & 65.50  & 47.28 & 45.75 & 74.43 & \cellcolor{col11}72.52 & \cellcolor{col33}58.78 & \cellcolor{col11}71.80  & \cellcolor{col33}59.17 \\
    
    & FFT & {CVPR'25}   & {\color{black}val-free} & \cellcolor{col11}72.32 & \cellcolor{col33}59.33 & 64.85 & 48.10 & 49.03 & 75.33  & \cellcolor{col11}71.18 & \cellcolor{col33}59.33 & \cellcolor{col11}70.35 & \cellcolor{col33}59.21 \\

    & FFT  & {CVPR'25}   & {\color{gray}ID test acc.} & \cellcolor{col11}72.82 & \cellcolor{col33}60.66 & 65.45 & 48.76 & 51.05 & 77.39 & \cellcolor{col11}72.00 & \cellcolor{col33}60.88 & \cellcolor{col11}71.02 & \cellcolor{col33}60.24 \\
    
    & SWAT & {CVPR'25} & {\color{black}val-free} & \cellcolor{col11}73.48 & \cellcolor{col33}63.35 & 66.51 & \textbf{54.86} & 50.86 & \textbf{81.20} & \cellcolor{col11}72.47 & \cellcolor{col33}\underline{63.15} & \cellcolor{col11}72.03 & \cellcolor{col33}\underline{63.02} \\
    
    & SWAT & {CVPR'25}   & {\color{gray}ID test acc.} & \cellcolor{col11}73.48 & \cellcolor{col33}63.35 & 66.51 & \textbf{54.86} & 50.86 & \textbf{81.20} & \cellcolor{col11}\underline{72.57} & \cellcolor{col33}63.01 & \cellcolor{col11}\underline{72.14} & \cellcolor{col33}62.86 \\

    & \textbf{VEST}  & {\textbf{ours}} & gF1 score  &
    \cellcolor{col11}\textbf{75.39} & \cellcolor{col33}\textbf{64.24} & \textbf{68.46} & \underline{54.55} & 52.93 & \underline{81.02} &  \cellcolor{col11}\textbf{74.36}  &  \cellcolor{col33}\textbf{64.06} &  \cellcolor{col11}\textbf{73.26} &  \cellcolor{col33}\textbf{63.46} \\
    \bottomrule
    \end{tabular}}

\vspace{-1mm}
\end{table}
}

\textbf{Compared Methods.}
We compare representative VLM adaptation methods categorized into four groups: prompt tuning~\citep{jia2022visual, pratt2023does, zhou2022learning, zhou2022conditional, khattak2023maple, Khattak2024ProText, 
wang2025attention, liang2026prompt}, adapter learning~\citep{zhang2022tip, gao2023clip, yu2023task, zhang2024dual, hu2022lora, zanella2024low}, linear probing~\citep{radford2021learning, wortsman2022robust, lin2023multimodality, shi2024lca, clap24}, and finetuning~\citep{wortsman2022robust, goyal2023finetune, liu2025few}. 
Fig.~\ref{fig:vlm_adapt} depicts their diagrams.
For fair comparison, we run their open-source code to train on the same ID training data.
Most of these methods rely on a validation set or even the test set for hyperparameter tuning and checkpoint selection, as also reported by~\citep{clap24, liu2025few}.
For them, we use the ID test set for tuning, although this helps them report over-optimistic results.
Moreover, CLAP~\citep{clap24}, few-shot finetuning (FFT) and SWAT~\citep{liu2025few} are validation-free methods. 
We adopt their reported hyperparameters and train them without validation.
Moreover, other than our VEST, SWAT also leverages external data to facilitate FSR. 
In experiments, we use the same retrieved data for both.
We include our code in the supplement and will release it to the public.
Refer to Appendix \S\ref{sec:compared_methods} for details of these methods.

{\bf Implementation Details.}
We implement all the methods with PyTorch and train on 
one NVIDIA RTX 4090 GPU (24GB).
For all the methods, we apply the same data augmentation: random crop and horizontal flip.
In training,
we use the AdamW optimizer and a cosine-annealing learning rate scheduler.
For our method, we set the learning rates to 1e-6 for the backbone and 1e-3 for the classifier
and batch size to 64.
We finetune the VLM for 10 epochs per stage and use our validation strategy for checkpoint selection.

\begin{table}[t]
{
\setlength{\tabcolsep}{0.42em} 
\begin{minipage}{0.45\textwidth}
  \centering
  \scalebox{0.68}{
    \begin{tabular}{ccclll}
    \toprule
    \multicolumn{2}{c}{\bf Stage-1} & {\bf Stage-2} & \multicolumn{2}{c}{accuracy} & \multirow{2}[2]{*}{\makecell{wall-clock \\ time ($h$)}}\\
    \cmidrule(lr){1-2}
    \cmidrule(lr){3-3}
    \cmidrule(lr){4-5}\
    tuning  & RA & tuning & \multicolumn{1}{l}{ID test}    &  \multicolumn{1}{l}{OOD test} \\
    \midrule
    F     &    \ding{53} &   \ding{53}   & 72.58 & 60.94 &  \color{gray}\ \ 0.5 \\
    F     & \checkmark     &    \ding{53}    & 72.89$^{\textcolor{Green}{+0.31}}$  & 63.28$^{\textcolor{Green}{+2.34}}$ &  \color{gray}\ \ 6.4 \\
    F     & \checkmark     &    F   & 74.49$^{\textcolor{Green}{+1.91}}$  & 63.58$^{\textcolor{Green}{+2.64}}$ &  \color{gray}\ \ 6.4 + 0.5 \\
    F     & \checkmark     &    P   &  74.74$^{\textcolor{Green}{+2.16}}$ & 64.09$^{\textcolor{Green}{+3.15}}$ &  \color{gray}\ \ 6.4 + 0.4\\

    \midrule
    P     &    \ding{53}    &   \ding{53}    & 73.35 & 61.80 & \color{gray}\ \ 0.4 \\
     P    & \checkmark     &    \ding{53}  & 73.56$^{\textcolor{Green}{+0.21}}$ & 63.60$^{\textcolor{Green}{+1.80}}$ & \color{gray}\ \ 3.5 \\
      P   & \checkmark     &   F   & 75.38$^{\textcolor{Green}{+2.03}}$ & 64.07$^{\textcolor{Green}{+2.07}}$ & \color{gray}\ \ 3.5 + 0.5 \\
       \cellcolor{col33}P   & \cellcolor{col33}\checkmark     &   \cellcolor{col33}P   & \cellcolor{col33}\textbf{75.39}$^{\textcolor{Green}{+2.04}}$ & \cellcolor{col33}\textbf{64.24}$^{\textcolor{Green}{+2.44}}$ & \cellcolor{col33}\color{gray}\ \ 3.5 + 0.4 \\

    \bottomrule
    \end{tabular}}
    \vspace{3mm}
\end{minipage}}
\hfill
\begin{minipage}{0.54\textwidth}
    \caption{\small
    {\bf Ablation study on finetuning strategies.} 
    We compare different combinations of stage-1 and stage-2 finetuning strategies, with or without RA, and 
    with either full finetuning (``F'') or partial finetuning (``P'').
    All models use our gF1 for checkpoint selection, with wall-clock time reported for 10 epochs of finetuning. 
    Incorporating RA in stage-1 yields notable gains in OOD accuracy, while subsequent stage-2 further improves both ID and OOD performance. 
    VEST, highlighted in {\setlength{\fboxsep}{1pt}\colorbox{col33}{the last row}}, achieves the best ID and OOD accuracy with decent training efficiency.
    }
    \label{tab:components}
\end{minipage}
\vspace{-4mm}
\end{table}

\begin{table}[t]
{
\setlength{\tabcolsep}{0.81em} 
\begin{minipage}{0.45\textwidth}
  \centering
  \scalebox{0.7}{
    \begin{tabular}{cccll}
    \toprule
    \multicolumn{2}{c}{\bf Stage-1} & {\bf Stage-2} & \multicolumn{2}{c}{accuracy} \\
    \cmidrule(lr){1-2}
    \cmidrule(lr){3-3}
    \cmidrule(lr){4-5}\
    tuning  & RA & tuning & \multicolumn{1}{l}{ID test}    &  \multicolumn{1}{l}{OOD test} \\
    \midrule

    \rowcolor{lightlightgrey}
    F     &    \ding{53} &   \ding{53}   & 72.32 & 59.33  \\
    F     &    \ding{53} &   \ding{53}   & 72.58$^{\textcolor{Green}{+0.26}}$ & 60.94$^{\textcolor{Green}{+1.61}}$  \\
    
    \rowcolor{lightlightgrey}
    F     & \checkmark     &    \ding{53}    & 71.43  & 59.51\\
    F     & \checkmark     &    \ding{53}    & 72.89$^{\textcolor{Green}{+1.46}}$  & 63.28$^{\textcolor{Green}{+3.77}}$  \\
    
    \rowcolor{lightlightgrey}
    F     & \checkmark     &    F   & 73.52 & 59.92  \\
    F     & \checkmark     &    F   & 74.49$^{\textcolor{Green}{+0.97}}$  & 63.58$^{\textcolor{Green}{+3.66}}$  \\

    \bottomrule
    \end{tabular}}
    \vspace{3mm}
\end{minipage}}
\hfill
\begin{minipage}{0.54\textwidth}
    \caption{\small
    {\bf Validating gF1 by comparing against the validation-free strategy} (in gray-shaded rows),
    which defaults to the final checkpoint at each training stage.
    We include settings of within/without RA, and with/without stage-2 training.
    As the validation-free strategy lacks a mechanism to determine the layer depth to finetune,
    we adopt full finetuning (F) in this comparison.
    In all settings, gF1 improves over the validation-free counterpart.
    }
    \label{tab:gF1_checkpoint_selection}
\end{minipage}
\vspace{-4mm}
\end{table}

\begin{table}[t]
\begin{minipage}{0.4\textwidth}
{
\setlength{\tabcolsep}{0.7em} 
\scalebox{0.7}{
\begin{tabular}{lcccc}
    \toprule
    & ckpt & ID test set & OOD test set \\
   \midrule
   w/o val & 50 &  72.32 & 59.33 \\
   \midrule
    rand-25 & 4 & 72.75 & 60.82 \\
    rand-50 & 3 & 72.58 & 60.94 \\
    rand-100 & 3 & 72.58 & 60.94 \\
    high-quality & 4 & 72.75 & 60.82 \\
    low-quality & 3 & 72.58 & 60.94 \\
    \bottomrule
\end{tabular}}
\vspace{3mm}
}
\end{minipage}
\hfill
\begin{minipage}{0.58\textwidth}
    \caption{\small
     {\bf Robustness of gF1 to validation data quality and quantity.}
     We construct validation sets of varying quality (top vs. bottom 50 retrieved images by similarities between retrieved image captions and target class names), and quantity (random 25, 50, and 100 images per class). Using these sets for FFT checkpoint selection demonstrates that gF1 is highly robust to retrieved data variations used for validation.
      }
    \label{tab:gF1_sensitive}
\end{minipage}
\vspace{-4mm}
\end{table}

\subsection{Experimental Results}

{\bf Benchmark results.}
\Cref{tab:baseline}  compares our VEST and various VLM adaptation methods.
Notably, for the methods that use a validation set of ID data for hyperparameter tuning and checkpoint selection, we do the same here, reporting their over-optimistic accuracies and marking their validation method as ``{\color{gray}ID test acc}. '''
Moreover, a few methods are validation-free, including CLAP, FFT, and SWAT; both SWAT and our VEST adopt RA, exploiting the same set of retrieved data in experiments.
We also report the results of FFT and SWAT using the ID test set for validation to establish an artificial ``upper bound'' on their performance.
Under this setup,
FFT achieves slightly higher accuracy (ref. the fourth last row);
SWAT yields a marginal improvement on the ID test data but a marginal decrease on the OOD test data (ref. the second last row).
Although SWAT remains stable even without validation \citep{liu2025few},
the validation by our gF1 enables extensive hyperparameter tuning, ultimately allowing our method VEST to achieve state-of-the-art performance on both ID and OOD test data.

{\bf Validation-enabled layer selection.}
Recall that our validation strategy allows layer selection for partial finetuning (PFT), which resoundingly outperforms the validation-free few-shot finetuning (FFT) method~\citep{liu2025few} and linear probing.
Fig.~\ref{fig:layer-selection} in \S\ref{ssec:layer_selection} has reported the results.
Hereafter in experiments,  PFT refers to finetuning the top-4 blocks of VLM's visual encoder.

{\bf Stage-wise finetuning with RA.}
In \Cref{tab:components}, we ablate RA, PFT (enabled by our validation strategy), stage-wise tuning. 
Regardless of adopting RA, 
PFT consistently outperforms FFT, corroborating that our validation strategy helps determine what parameters to finetune to achieve better test-time generalization.
Moreover, RA substantially improves OOD accuracy while it provides limited gains in ID accuracy for both FFT and PFT. 
By adopting another stage of finetuning using the few-shot ID training data only,
we obtain remarkable accuracy gains on both ID and OOD test data.
We further evaluate the effect of gF1 for checkpoint selection in~\Cref{tab:gF1_checkpoint_selection}.
Applying gF1 for checkpoint selection substantially improves ID and OOD test accuracy over validation-free methods across different finetuning strategies.
Lastly, VEST has decent computation efficiency.

{\bf Robustness of gF1 to the quality and quantity of retrieved data.} 
Table~\ref {tab:gF1_sensitive} reports the results of selected checkpoints by  gF1  using different sets of retrieved data with varying quality and quantity.
Results show that gF1 robustly selects checkpoints that perform well on both ID and OOD test data.
We provide more robustness analyses  regarding more aspects in Appendix \S\ref{sec:more_gF1_results}.

{\bf Cross-dataset validation.}
To further validate our validation strategy,
we simulate a distributionally shifted validation set by using each one of the four ImageNet OOD test sets.
We perform validation for checkpoint selection through few-shot finetuning (FFT).
\Cref{tab:cross_evaluation} reports the results, showing that, regardless of the OOD data used for validation, our gF1 consistently selects checkpoints that resoundingly outperform the recent validation-free FFT method~\citep{liu2025few}.

{\bf Validation for finetuning DINOv2.}
\Cref{tab:baseline_dinov2} shows that our gF1 validation strategy helps various methods when employing the more powerful DINOv2 backbone \citep{oquab2024dinov}.
Refer to the table caption and Appendix \S\ref{sec:more_gF1_results} for more details.

{
\begin{table}[t]
\begin{minipage}{0.48\textwidth}
  \centering
    \vspace{-1.5mm}
  \caption{\small
  {\bf Cross-dataset validation.}
  With each one of the four ImageNet OOD test sets as the validation set, we apply our validation strategy for checkpoint selection in few-shot finetuning (FFT) \citep{liu2025few}.
  We report the accuracies of the selected checkpoint on the ID test set, the other three OOD test sets, as well as {\setlength{\fboxsep}{1pt}\colorbox{lightlightgrey}{the val-set itself}}.
  Compared to the validation-free FFT method (i.e., the first row), our gF1 score can select superior checkpoints regardless of the OOD dataset used for validation.
  }
  \vspace{-3.5mm}
  \scalebox{0.58}{
    \begin{tabular}{l l l l l l l l}
    \toprule
    \multirow{2}[3]{*}{val-set} & \multicolumn{1}{c}{ID test}    & \multicolumn{4}{c}{OOD test} \\
    \cmidrule(lr){2-2}
    \cmidrule(lr){3-6}
    & IN  & IN-V2 & IN-S & IN-A & IN-R \\
    \midrule
    w/o val & 72.32 & 64.85 & 48.10 & 49.03 & 75.33 \\
    \midrule
    IN-V2 & 72.75$^{\textcolor{Green}{+0.43}}$ 
        & \cellcolor{lightlightgrey}65.54    & 49.02$^{\textcolor{Green}{+0.92}}$ & 51.04$^{\textcolor{Green}{+2.01}}$ & 77.69$^{\textcolor{Green}{+2.36}}$ \\
    
    IN-S & 72.58$^{\textcolor{Green}{+0.26}}$
        & 65.54$^{\textcolor{Green}{+0.69}}$ & \cellcolor{lightlightgrey}49.39    & 50.72$^{\textcolor{Green}{+1.69}}$ & 78.12$^{\textcolor{Green}{+2.79}}$ \\
    
    IN-A & 72.68$^{\textcolor{Green}{+0.36}}$ 
        & 65.46$^{\textcolor{Green}{+0.61}}$ & 48.72$^{\textcolor{Green}{+0.62}}$ & \cellcolor{lightlightgrey}51.15     & 77.09$^{\textcolor{Green}{+1.76}}$ \\
    
    IN-R & 72.58$^{\textcolor{Green}{+0.26}}$ 
        & 65.54$^{\textcolor{Green}{+0.69}}$ & 49.39$^{\textcolor{Green}{+1.29}}$ & 50.72$^{\textcolor{Green}{+1.69}}$ & \cellcolor{lightlightgrey}78.12 \\
    
    \bottomrule
    \end{tabular}}
  \label{tab:cross_evaluation}
\end{minipage}
\hfill
\begin{minipage}{.48\textwidth}
    \centering
    \vspace{-1.5mm}
  \caption{\small
  When employing a more powerful backbone DINOv2, 
  FFT, PFT, and VEST also benefit from adopting gF1 for validation   
  FFT finetunes DINOv2 for 150 epochs in the 16-shot setting. 
  The checkpoint selected by gF1
  outperforms the FFT's default checkpoint (i.e., the last one).
  Moreover, gF1 supports layer selection for partial finetuning (PFT), which yields further improvements. 
  Lastly, our VEST  performs the best with DINOv2,
  remarkably better than the CLIP visual encoder (cf. Table~\ref{tab:baseline}).
  }
  \vspace{-2.5mm}
  \setlength{\tabcolsep}{0.4em}
  \scalebox{0.65}{
    \begin{tabular}{llcccccc}
    \toprule
         \multirow{2}[2]{*}{methods} & \multirow{2}[2]{*}{validation} & ID test & \multicolumn{5}{c}{OOD test} \\
    \cmidrule(lr){3-3}
    \cmidrule(lr){4-8}
     &  & \cellcolor{col11}IN    & \cellcolor{col33}avg.  & IN-V2 & IN-S  & IN-A  & IN-R \\
    \midrule
    FFT   & {\color{black}val-free} &  \cellcolor{col11}77.01	&  \cellcolor{col33}55.56   &  68.25    &  47.88   &   48.44 &    57.66  \\
    FFT   & {\color{black}gF1 score} &  \cellcolor{col11}77.73 &  \cellcolor{col33}58.50  &   69.60    &   50.26   &   53.20  &   60.92   \\
    PFT   & {\color{black}gF1 score} &   \cellcolor{col11}77.15    &  \cellcolor{col33}60.62  &  69.44  &   52.85  &   52.76   &   67.44  \\
    {\bf VEST}  & {\color{black}gF1 score} &   \cellcolor{col11}\textbf{79.37}    &  \cellcolor{col33}\textbf{67.49}     &   \textbf{72.33}    &   \textbf{60.17}    &   \textbf{58.68}    &   \textbf{78.77} \\
    \bottomrule
    \end{tabular}}
  \label{tab:baseline_dinov2}
\end{minipage}
\vspace{-3mm}
\end{table}
}

\section{Conclusions}

This work addresses the critical need for 
robust generalization in few-shot recognition (FSR) via Vision-Language Model adaptation, 
aiming for high accuracy across both in-distribution (ID) and out-of-distribution (OOD) test data.
To mitigate the core challenge of data scarcity, 
not only the limited training data but also a complete lack of validation data,
we exploit retrieved data for validation and develop a novel and effective validation metric, gF1 score, enabling hyperparameter tuning toward improved generalization.
We further leverage this open source to adapt VLMs in a stage-wise manner, deriving our final method,
Validation-Enabled Stage-wise Tuning (VEST).
Extensive experiments on five benchmarks show that VEST resoundingly outperforms existing approaches.

\section*{Acknowledgments}
The authors thank anonymous reviewers for their constructive feedback, which  substantially improved the manuscript. 
SK gratefully acknowledges support from 
the CK Foundation and the Plato Initiative.
This work was supported by 
the Science and Technology Development Fund of Macau (0058/2025/RIA2, 0067/2024/ITP2),  and
the University of Macau (MYRG-CRG2026-00038-FIC, MYRG-GRG2026-00236-FIC, SRG2023-00044-FST).

{
\small
\bibliography{arxiv}
\bibliographystyle{plainnat}
}

\appendix
\clearpage

{
\begin{center}
\Large
\textbf{Enabling Validation for Robust Few-Shot Recognition}\\
(Appendix)
\end{center}
}

\renewcommand{\thesection}{\Alph{section}}
\renewcommand{\theHsection}{\Alph{section}}
\setcounter{section}{0}

The appendix is organized as follows: 

\begin{itemize}

\item 
{\bf Section \ref{sec:proof}} presents theorems and proofs.

\item {\bf Section \ref{sec:datasets}}  describes datasets used in our experiments and shows more examples of training and testing data.

\item {\bf Section \ref{sec:retrieval}}  provides the implementation details of open data retrieval.

\item {\bf Section \ref{sec:hyperparams}} presents hyperparameters used in our work.

\item {\bf Section \ref{sec:more_gF1_results}} provides more results of our validation strategy for checkpoint selection and hyperparameter tuning.

\item {\bf Section \ref{sec:compared_methods}} provides more details of the compared methods.

\item {\bf Section \ref{sec:other_datasets}} provides the results on other datasets.

\item {\bf Section \ref{sec:more_discussions}} discusses limitations and future work.

\item {\bf Section \ref{sec:code}}  provides code to reproduce our results.

\end{itemize}

\section{Theorems and Proofs}
\label{sec:proof}

\subsection{Theorem 1 and Proof}

We copy Theorem 1 from the main paper:
\\
\\
\textbf{Theorem 1}
\textit{
Let $\mathcal{F} \subset \mathbb{R}_{>0}^n$ be a feasible set of strictly positive objective vectors. If $\mathbf{x}^* \in \mathcal{F}$ is a global maximizer of the harmonic mean over $\mathcal{F}$, i.e.,
\[
    H(\mathbf{x}^*) \ge H(\mathbf{x}) \quad \forall \mathbf{x} \in \mathcal{F},
\]
then $\mathbf{x}^*$ is Pareto-optimal.
}

To prove this,
let us introduce the following definitions.
\begin{definition}[Pareto Dominance and Optimality]
Let $\mathcal{F} \subset \mathbb{R}_{>0}^n$ be a feasible set of strictly positive objective vectors. A vector $\mathbf{y} \in \mathcal{F}$ \emph{Pareto-dominates} $\mathbf{x} \in \mathcal{F}$ (denoted $\mathbf{y} \succ \mathbf{x}$) if:
\begin{enumerate}
    \item $y_i \ge x_i$ for all $i \in \{1, \dots, n\}$, and
    \item $y_k > x_k$ for at least one index $k \in \{1, \dots, n\}$.
\end{enumerate}
A point $\mathbf{x}^* \in \mathcal{F}$ is \emph{Pareto-optimal} (lies on the Pareto front) if there exists no $\mathbf{y} \in \mathcal{F}$ such that $\mathbf{y} \succ \mathbf{x}^*$.
\end{definition}

\begin{definition}[Harmonic Mean]
For any strictly positive vector $\mathbf{x} = (x_1, \dots, x_n) \in \mathbb{R}_{>0}^n$, the harmonic mean $H: \mathbb{R}_{>0}^n \to \mathbb{R}_{>0}$ is defined as
\begin{equation}
    H(\mathbf{x}) = \frac{n}{\sum_{i=1}^n \frac{1}{x_i}}.
\end{equation}
\end{definition}

\textbf{Proof of Theorem 1:}
\\
We proceed by contradiction. Suppose that $\mathbf{x}^* \in \mathcal{F}$ maximizes $H(\mathbf{x})$ over $\mathcal{F}$, yet $\mathbf{x}^*$ is not Pareto-optimal.
By the definition of Pareto optimality, there must exist a feasible vector $\mathbf{y} \in \mathcal{F}$ such that $\mathbf{y} \succ \mathbf{x}^*$. Consequently, $y_i \ge x_i^*$ for all $i \in \{1, \dots, n\}$, and there exists at least one index $k \in \{1, \dots, n\}$ such that $y_k > x_k^*$.

Since $\mathcal{F} \subset \mathbb{R}_{>0}^n$, every component satisfies $y_i > 0$ and $x_i^* > 0$. Taking reciprocals reverses the inequalities:
\begin{align}
    \frac{1}{y_i} &\le \frac{1}{x_i^*} \quad \forall i \in \{1, \dots, n\}, \\
    \frac{1}{y_k} &< \frac{1}{x_k^*} \quad \text{for index } k.
\end{align}
Summing these inequalities across all $i = 1, \dots, n$, the strict inequality at index $k$ yields
\begin{equation}
    \sum_{i=1}^n \frac{1}{y_i} < \sum_{i=1}^n \frac{1}{x_i^*}.
\end{equation}
Because both sums are strictly positive and $n \ge 1$, dividing $n$ by the strictly smaller denominator produces a strictly larger value:
\begin{equation}
    H(\mathbf{y}) = \frac{n}{\sum_{i=1}^n \frac{1}{y_i}} > \frac{n}{\sum_{i=1}^n \frac{1}{x_i^*}} = H(\mathbf{x}^*).
\end{equation}
This implies $H(\mathbf{y}) > H(\mathbf{x}^*)$ for some $\mathbf{y} \in \mathcal{F}$, which contradicts the assumption that $\mathbf{x}^*$ is a global maximizer of $H(\mathbf{x})$ over $\mathcal{F}$. Thus, no such dominating point $\mathbf{y}$ can exist, and $\mathbf{x}^*$ is Pareto-optimal.

\subsection{Theorem 2 and Proof}

We copy Theorem 2 from the main paper:
\\
\\
\textbf{Theorem 2}
\textit{
Let $\mathcal{F} \subset \mathbb{R}_{>0}^n$ be a feasible set of strictly positive objective vectors.
Let $\{\mathcal{F}_i\}_{i \in \mathcal{I}}$ be an arbitrary collection of subsets of $\mathcal{F}$ indexed by a set $\mathcal{I}$,
and let $\mathcal{F}  = \bigcup_{i \in \mathcal{I}} \mathcal{F}_i$ be their union.
Then the Pareto front of $\mathcal{F}$ is a subset of the union of the Pareto front of $\mathcal{F}_i$:
$\mathcal{P}\left(\bigcup_{i \in \mathcal{I}} \mathcal{F}_i\right) \subseteq \bigcup_{i \in \mathcal{I}} \mathcal{P}(\mathcal{F}_i)$.
}

\textbf{Proof of Theorem 2:}\\
Let $\x^* \in \mathcal{P}(\mathcal{F})$ be an arbitrary Pareto-optimal point in the union $\mathcal{F} = \bigcup_{i \in \mathcal{I}} \mathcal{F}_i$.
If $\mathcal{P}(\mathcal{\mathcal{F}}) = \emptyset$, 
the inclusion holds trivially.

Since $\x^* \in \mathcal{P}(\mathcal{F}) \subseteq \mathcal{F}$, 
by definition of the union there exists at least one index $j \in \mathcal{I}$ such that $\x^* \in \mathcal{F}_j$.
We claim that $\x^* \in \mathcal{P}(\mathcal{F}_j)$.

Suppose, for the sake of contradiction, that $\x^* \notin \mathcal{P}(\mathcal{F}_j)$.
Because $\x^* \in \mathcal{F}_j$, 
this implies that $\x^*$ is dominated within $\mathcal{F}_j$;
that is, there exists a point $\y \in \mathcal{F}_j$ such that
\[
\y \succ \x^*.
\]
However, since $\mathcal{F}_j \subseteq \bigcup_{i \in \mathcal{I}} \mathcal{F}_i = \mathcal{F}$, 
it follows that $\y \in \mathcal{F}$. 
Consequently, there exists $\y \in \mathcal{F}$ such that $\y \succ \x^*$, which contradicts the premise that $\x^* \in \mathcal{P}(\mathcal{F})$.

Thus, the assumption is false, and we conclude that $\x^* \in \mathcal{P}(\mathcal{F}_j)$. Since $j \in \mathcal{I}$, 
it follows that
\[
\x^* \in \bigcup_{i \in \mathcal{I}} \mathcal{P}(\mathcal{F}_i).
\]
Because $\x^* \in \mathcal{P}(\mathcal{F})$ was chosen arbitrarily, we obtain $\mathcal{P}\left(\bigcup_{i \in \mathcal{I}} \mathcal{F}_i\right) \subseteq \bigcup_{i \in \mathcal{I}} \mathcal{P}(\mathcal{F}_i)$.

\textbf{Corollary 1}
\textit{Suppose that $\mathcal{I}$ is non-empty finite index set
and each subset $\mathcal{F}_i$ ($i\in\mathcal{I}$) is non-empty and finite (or compact). 
Then 
the global Pareto front $\mathcal{P}\left(\bigcup_{i \in \mathcal{I}} \mathcal{F}_i\right)$ can be recovered directly from the union of local Pareto fronts: 
$\mathcal{P}\left(\bigcup_{i \in \mathcal{I}} \mathcal{F}_i\right) = \mathcal{P}\left(\bigcup_{i \in \mathcal{I}} \mathcal{P}(\mathcal{F}_i)\right)$.
}

\section{Benchmark Datasets}
\label{sec:datasets}

Table~\ref{tab:datasets} summarizes the statistics of all datasets used in our experiments. 
For few-shot recognition, we randomly sample $\{4, 8, 16\}$ images from the ID training set to construct our training set and evaluate methods on both ID and OOD test sets.
Fig.~\ref{fig:other_datasets_supp} visualizes the image feature distribution of examples sampled from the ImageNet training and testing sets,
highlighting the distributional shifts in each benchmark. 
Results on ImageNet-based OOD benchmark are reported in \Cref{tab:baseline}, while results on the remaining datasets are presented in \cref{sec:other_datasets}.
Below, we provide a detailed description of each dataset.

\begin{itemize}
    \item 
    \textbf{ImageNet-based OOD benchmark} is a large-scale benchmark built upon ImageNet~\citep{deng2009imagenet}, which contains one million images across 1,000 common classes. It uses ImageNet as the ID dataset and its four variants as OOD test sets: ImageNet-V2~\citep{recht2019imagenet}, ImageNet-S~\citep{wang2019learning}, ImageNet-A~\citep{hendrycks2021natural}, and ImageNet-R~\citep{hendrycks2021many}. 
    Specifically, ImageNet-V2 contains later collected images for all ImageNet classes, while ImageNet-S provides sketch-style images for the same 1,000 classes. ImageNet-A contains adversarial images for a subset of 200 ImageNet classes, and ImageNet-R includes artistic renditions for another subset of 200 classes.

    \item
    \textbf{Living17, and Entity30} are part of the BREEDS benchmark~\citep{santurkar2020breeds}, which are derived from ImageNet and designed specifically to evaluate subpopulation shift. Their ID and OOD sets contain different breeds or subtypes of the same category. Specifically,
    Living17 consists of images from 17 animal classes, while Entity30 comprises 30 entity classes, such as dish and vegetable.

    \item 
    \textbf{fMoW}~\citep{christie2018functional} contains satellite images spanning 62 building and land-use categories across six geographic regions: Africa, Americas, Oceania, Asia, Europe, and Other.
    Following the setting in \citep{kumar2022finetuning}, we use images from the Americas as ID data and images from Africa and Europe as OOD data, with the selected regions covering 60 classes in total.

    \item 
    \textbf{DomainNet}~\citep{peng2019moment} is a widely used domain adaptation benchmark, which consists of 345 categories across six distinct domains: Clipart, Infograph, Painting, Quickdraw, Real, and Sketch. Following the setting in \citep{kumar2022finetuning}, we use sketch images as ID data, and images of the same categories from the remaining five domains as OOD data. 
    
\end{itemize}

{
\begin{table}[b]
\centering
\small

\begin{minipage}{0.42\textwidth}
    \vspace{-3mm}
    \setlength{\tabcolsep}{1.5em}
    \scalebox{0.8}{
    \begin{tabular}{lrr}
    \toprule 
    dataset & \# classes & \# test data \\
    \midrule
    ImageNet & 1,000 & 50,000  \\
    ImageNet-V2 & 1,000  & 10,000  \\
    ImageNet-S &1,000 & 50,000 \\
    ImageNet-A &200 & 7,500  \\
    ImageNet-R &200 & 30,000 \\

    \midrule

    Living17 & 17 & 3,400 \\
    Entity30 & 30 & 12,000 \\
    fMoW & 60 & 16,126 \\
    DomainNet & 345 & 176,743 \\
    
    \bottomrule
    \end{tabular}}
\end{minipage}
\hspace{\fill}
\begin{minipage}{0.56\textwidth}
    \caption{\small \textbf{Statistics of datasets for evaluating OOD generalization.}
    We list the number of classes and images in the test sets for all the datasets we used.  
    Following the few-shot OOD evaluation protocol, 
    we randomly sample $\{4, 8, 16\}$ images from the ID training set to adapt the VLM (referred to as \emph{our training set}) and evaluate the model's performance on both ID and OOD test sets.
    Notably, we do not exploit a validation set. 
    Instead, we propose a novel validation strategy by exploiting retrieved open data.
    }
    \label{tab:datasets}
\end{minipage}
\vspace{-3mm}
\end{table}
}

\begin{figure}[t]
    \begin{minipage}{0.50\textwidth}
        \centering
    \includegraphics[width=0.88\linewidth, clip=true, trim = 4mm 4mm 0mm 0mm]{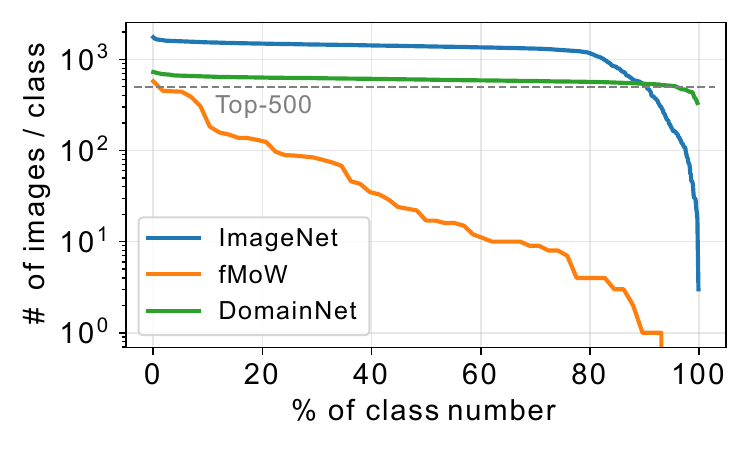}
    \vspace{-2mm}
    \caption{\small
    {\bf The retrieved data from LAION-400M follows an imbalanced distribution.} We retrieve relevant samples via string matching, subsequently selecting the 500 most similar images per class (10 images per class for fMoW) by text-to-text feature similarity for training, and randomly sampling 50 images per class (10 for fMoW) from the remaining data for validation. 
    }
    \label{fig:distribution_suppl}
    \end{minipage}
    \hspace{\fill}
    \begin{minipage}{0.46\textwidth}
    \centering
    \includegraphics[width=\linewidth]{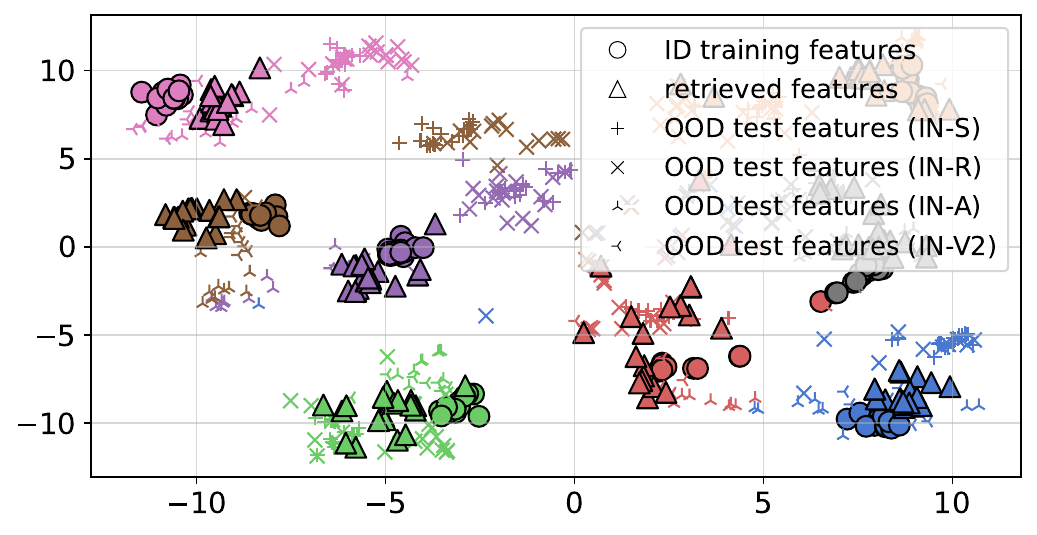}
    \vspace{-4mm}
    \caption{\small
    {\bf t-SNE visualization} of features of ID training data, OOD test data, and retrieved data from eight randomly selected classes. All features are extracted from the pretrained CLIP model. While the three data sources have clear distribution shifts, the retrieved data bridges distribution gaps between the ID training data and OOD test data.
    }
    \label{fig:tsne_for_RA}
    \end{minipage}
\vspace{-0mm}    
\end{figure}

\section{Details of Open Data Retrieval}
\label{sec:retrieval}

To find task-relevant data, early methods rely on feature similarities between target class names and pretraining images/texts~\citep{liu2023learning, wallingford2023neural}.
This requires downloading, hosting, and computing over hundreds of millions of images and texts.
In contrast, recent works use string matching to find relevant image captions first and then retrieve corresponding images~\citep{parashar2024neglected, liu2025few}, demonstrating
significantly enhanced efficiency and a capability to retrieve more visually diverse images.

Following this line of work~\citep{liu2023learning, liu2025few},
we retrieve images from LAION-400M~\citep{laion400m} via string matching.
As illustrated in Fig.~\ref{fig:distribution_suppl}, 
the retrieved data follows a long-tail distribution,
an imbalanced issue that exists in nature or available pretraining datasets such as LAION-400M and MetaCLIP~\citep{parashar2024neglected}.
For each class, 
we select 500 images to augment training data, except for fMoW, where only 10 images per class are used due to the limited number of retrieved samples for satellite images.
From the remaining data,
we randomly sample 50 images per class (10 for fMoW) to form a non-overlapping validation set.
Note that Living17 and Entity30 are derived from ImageNet and therefore share the same retrieved data.
It is possible that a class does not have sufficient data. 
In this case, we use all the retrieved images pertaining to this class for training and do not reserve any data for validation.

Retrieval examples for each dataset are shown in Fig.~\ref{fig:other_datasets_supp}. 
It is worth noting that the retrieved data have label noise and distributional shift relative to the ID training data \citep{liu2025few}.
Fig.~\ref{fig:tsne_for_RA} further visualizes the distributional shift in feature space.
Notably, the retrieved data lie closer to the ID training data in feature space than the OOD test data. 
They thus provide additional distributional coverage that is relevant to the target task while differing from the limited few-shot samples.
Incorporating these data facilitates the adaptation of decision boundaries and improves generalization to OOD data.

\section{Configuration of Hyperparameters}
\label{sec:hyperparams}

For our method, we employ the proposed gF1 score for both checkpoint selection and hyperparameter tuning. Following the practice in \citep{liu2025few, lin2023multimodality, parashar2024neglected, wortsman2022robust}, we initialize the classifier with the average text embedding of 80 OpenAI prompts for ImageNet. 
For other datasets, we adopt a simple prompt template: `a photo of a \{CLASS\}'.
The training pipeline includes two standard data augmentations: RandomResizedCrop and RandomHorizontalFlip. We adopt the AdamW optimizer~\citep{loshchilov2017decoupled} with a cosine annealing learning rate schedule and a batch size of 64.
For the CLIP model, the learning rate is set to 1e-6 for the backbone and 1e-3 for the classifier, except for fMoW, where we use 1e-5 and 1e-2, respectively.
For DINOv2, the learning rate is set to 1e-6 for the backbone and 1e-4 for the classifier.
A warmup phase of 18 iterations is applied, during which the learning rate increases linearly from 1e-8 to its initial value.
For partial finetuning (PFT), we train for 50 epochs with a weight decay of 0.1. For VEST, both stage 1 and stage 2 training are conducted for 10 epochs with a weight decay of 0.01.

For the compared methods, we directly adopt their reported hyperparameters and train them on the same few-shot training data for fair comparison.
To note, for the methods that require a validation set,
we use the ID test set in our experiments to help them with checkpoint selection, representing their upper-bound performance.

\begin{figure}
    \vspace{3mm}
    \centering
    \begin{minipage}{0.54\textwidth}
        \includegraphics[width=\linewidth, clip=true, trim = 0mm 0mm 0mm 0mm]{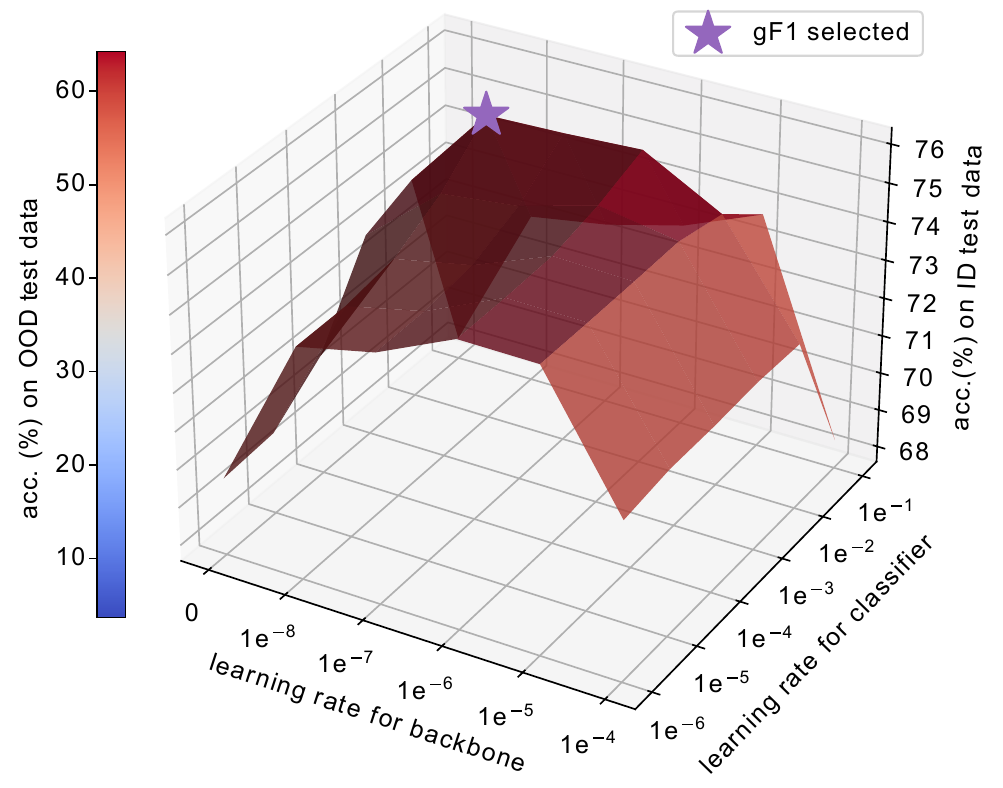}
    \end{minipage}
    \hfill
    \begin{minipage}{0.44\textwidth}
    \caption{\small
    {\bf Validation for tuning the learning rate.} 
    We apply the proposed validation strategy to tune the learning rate of the backbone and classifier for stage-2 finetuning of VEST. 
    For each candidate learning rate combination, we use gF1 to select a checkpoint and report its ID test accuracy (z-axis) and OOD test accuracy (surface color). 
    We observe strong performance on both ID and OOD test data across a broad range of backbone learning rates from 1e-6 to 1e-8 when the classifier learning rate is fixed at 1e-2. 
    Importantly, gF1 selects a learning rate combination within the high-performing region.
    }
    \label{fig:tuning_lr}
    \end{minipage}
\end{figure}

\section{Further analysis of our validation strategy}
\label{sec:more_gF1_results}

{\bf gF1 enables tuning more hyperparameters.}
We employ the AdamW optimizer~\citep{loshchilov2017decoupled} with separate base learning rates for the backbone and classifier in VEST. 
We perform a grid search over these two learning rates for VEST stage-2 using our proposed validation metric, gF1. 
Specifically, we search the backbone learning rate from \{0, 1e-8, 1e-7, 1e-6, 1e-5, 1e-4\} and the classifier learning rate from \{1e-6, 1e-5, 1e-4, 1e-3, 1e-2, 1e-1\}. 
Fig.~\ref{fig:tuning_lr} presents the resulting ID and OOD test accuracy for each learning rate combination. 
Our gF1 score selects a learning rate setting that yields better performance on both ID and OOD data, demonstrating its effectiveness.

{
\setlength{\tabcolsep}{0.9em} 
\begin{table}[t]
\centering
\caption{\small
{\bf Performance comparison between validation-free and validation by our gF1} across four representative finetuning methods. We apply gF1 for checkpoint selection and report the ID and OOD test accuracy, along with their average. Clearly, gF1 consistently improves the performance of different finetuning methods.}
\vspace{-3mm}
\label{tab:validation_for_other_methods}
\scalebox{0.85}{
\begin{tabular}{llcccccc}
\toprule
&  & \multicolumn{3}{c}{Validation-free} & \multicolumn{3}{c}{Validation by our gF1} \\
\cmidrule(lr){3-5} 
\cmidrule(lr){6-8}
& method & ID test & OOD test & \cellcolor{col3}avg. & ID test & OOD test & \cellcolor{col3}avg. \\
\midrule
PT & VPT{\scriptsize{shallow}} \citep{jia2022visual} & 72.60 & 57.46 & \cellcolor{col3}65.03 & 72.43  & 60.74  & \cellcolor{col3}66.59$^{\textcolor{Green}{+1.56}}$  \\
AL & LoRA~\citep{hu2022lora} & 71.53 & 56.55 & \cellcolor{col3}64.04 & 72.84  & 60.66  & \cellcolor{col3}66.75$^{\textcolor{Green}{+2.71}}$  \\
LP & CMLP~\cite{lin2023multimodality} & 73.16 & 59.32 & \cellcolor{col3}66.24 & 72.49  & 60.66  & \cellcolor{col3}66.58$^{\textcolor{Green}{+0.34}}$  \\
FT & FLYP~\citep{goyal2023finetune} & 73.89 & 58.15 & \cellcolor{col3}66.02 & 73.63  & 59.48  & \cellcolor{col3}66.56$^{\textcolor{Green}{+0.54}}$  \\
\bottomrule
\end{tabular}}
\vspace{-1mm}
\end{table}
}

{\bf gF1 is applicable to other VLM adaptation methods.} 
\Cref{tab:validation_for_other_methods} evaluates our gF1 for checkpoint selection on four representative VLM adaptation methods, covering the four categories in Fig.~\ref{fig:vlm_adapt}. Compared with the default validation-free setting that selects the final checkpoint, gF1 consistently selects better checkpoints and improves overall performance.

{\bf gF1 generalizes to the vision foundation model DINOv2.} 
To further evaluate the effectiveness of our validation method, we extend its application to the vision foundation model, specifically employing it for checkpoint and layer selection on a DINOv2 ViT-B/14 model with registers and a randomly initialized classifier~\citep{oquab2024dinov}.

\begin{figure*}[t]
\small
\centering
\begin{minipage}{0.56\textwidth}
    \centering
    \includegraphics[width=\linewidth]{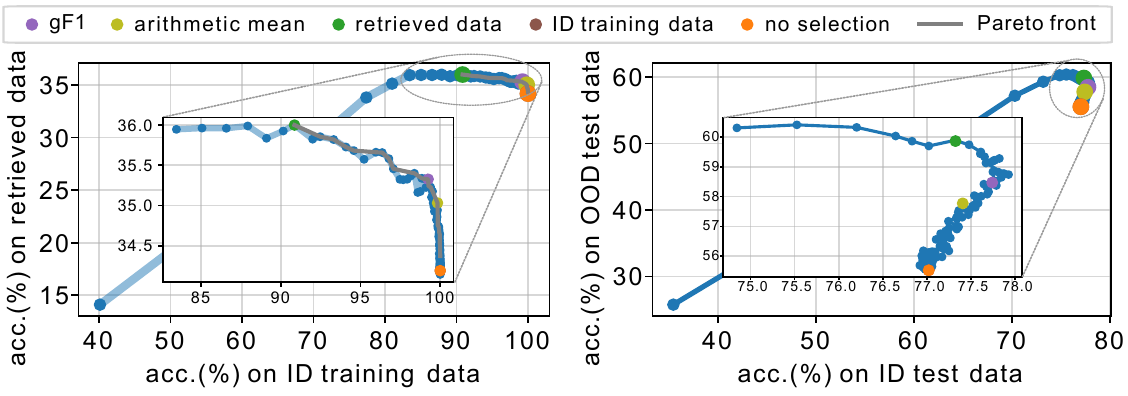}
\end{minipage}
\hfill
\begin{minipage}{0.42\textwidth}
\centering
{
\setlength{\tabcolsep}{.45em} 
\scalebox{0.62}{
\begin{tabular}{lccccccc}
\toprule
\multirow{2}[2]{*}{\makecell{validation \\ methods}} & \multirow{2}[2]{*}{ckpt} & ID test & \multicolumn{5}{c}{OOD test} \\
\cmidrule(lr){3-3}
\cmidrule(lr){4-8}
&  & \cellcolor{col11}IN & \cellcolor{col33}avg. & IN-V2 & IN-S & IN-A & IN-R \\
\midrule
no selection & 150 & \cellcolor{col11}77.01 & \cellcolor{col33}55.56 & 68.25 & 47.88 & 48.44 & 57.66 \\
ID train data & 150 & \cellcolor{col11}77.01 & \cellcolor{col33}55.56 & 68.25 & 47.88 & 48.44 & 57.66 \\
retrieved data & \ \ 11 & \cellcolor{col11}77.32 & \cellcolor{col33}\textbf{59.89} & \underline{69.48} & \textbf{51.50} & \textbf{54.12} & \textbf{64.45} \\
arithmetic mean & \ \ 47 & \cellcolor{col11}\underline{77.40} & \cellcolor{col33}57.80 & 69.14 & 49.82 & 52.00 & 60.24 \\
gF1 score & \ \ 32 &\cellcolor{col11}\textbf{77.73} & \cellcolor{col33}\underline{58.50} & \textbf{69.60} & \underline{50.26} & \underline{53.20} & \underline{60.92} \\
\bottomrule
\end{tabular}}}
\end{minipage}
\vspace{-3mm}
\caption{\small
{\bf Checkpoint selection on DINOv2.} We evaluate different validation strategies for checkpoint selection after fully finetuning a DINOv2 ViT-B/14 for 150 epochs.
 The right table presents the selected checkpoints (ref. ckpt) and their detailed performance on ID and OOD test data. The results demonstrate that our gF1 score generalizes effectively to vision foundation models, consistently selecting a checkpoint that achieves robust performance on both ID and OOD test data.
}
\label{fig:ckpt_selection_dinov2}
\end{figure*}

{
\setlength{\tabcolsep}{.9em} 
\begin{table}[t]
    \centering
    \caption{\small
    {\bf Layer selection on DINOv2.} We perform partial finetuning (PFT) on the top-$k$ layers of the vision model. We report top-1 accuracy on ID and OOD test data, along with the overall average, computed as the mean of ID accuracy and average OOD accuracy. A clear trade-off exists between ID and OOD test accuracy across different values of $k$. Our proposed validation method selects PFT top-1 layers, which provides a good balance between ID and OOD performance and remarkably outperforms linear probing and full finetuning.}
    \vspace{-3mm}
    \scalebox{0.8}{
    \begin{tabular}{lcccccccc}
    \toprule
    \multicolumn{1}{l}{\multirow{2}[4]{*}{finetuned layers}} & \multirow{2}[4]{*}{\makecell{gF1 \\ ($\times 10^{-2}$)}} & \multirow{2}[4]{*}{\makecell{overall \\ average}} & ID test   & \multicolumn{5}{c}{OOD test} \\
\cmidrule(lr){4-4}
\cmidrule(lr){5-9}
& &  & \cellcolor{col11}IN    & \cellcolor{col33}avg.  & IN-V2 & IN-S  & IN-A  & IN-R \\
    \midrule
    linear probing & 0.00 & \cellcolor{col3}68.29 & \cellcolor{col11}76.72 & \cellcolor{col33}59.86 & 68.82 & 52.45 & 51.20 & 66.97  \\
    \midrule
    
    top-1 & \textbf{0.43} & \cellcolor{col3}\textbf{68.89} & \cellcolor{col11}77.15 & \cellcolor{col33}\textbf{60.62} & 69.44 & 52.85 & 52.76	&  67.44  \\
    top-2 & 0.33& \cellcolor{col3}\underline{68.87} & \cellcolor{col11}77.66 & \cellcolor{col33}\underline{60.07} & 69.58	 & 52.35 & 53.07 & 65.29 \\
    top-3 & 0.00 & \cellcolor{col3}68.70 & \cellcolor{col11}77.80 & \cellcolor{col33}59.60 & 69.86 & 51.72 & 53.32 & 63.51   \\
    top-4 & 0.10 & \cellcolor{col3}68.62 & \cellcolor{col11}77.82 & \cellcolor{col33}59.42 & 69.97 & 51.48 & 53.49 & 62.73 \\
    top-5 & 0.08 & \cellcolor{col3}68.58 & \cellcolor{col11}77.97 & \cellcolor{col33}59.19 & 69.85 & 51.22 & 53.48 & 62.22 \\
    top-6 & 0.09 & \cellcolor{col3}68.56 & \cellcolor{col11}78.01 & \cellcolor{col33}59.10 & 69.79	& 50.96 & 53.60 & 62.04  \\
    top-7 & 0.04 & \cellcolor{col3}68.52 & \cellcolor{col11}\textbf{78.07} & \cellcolor{col33}58.97 & 69.63 & 50.71 & 53.83 & 61.70 \\
    top-8 & 0.00 & \cellcolor{col3}68.49 & \cellcolor{col11}\underline{78.02} & \cellcolor{col33}58.95 & 69.70 & 50.60 & 53.84 & 61.66 \\
    top-9 & 0.02 & \cellcolor{col3}68.50 & \cellcolor{col11}\underline{78.02} & \cellcolor{col33}58.98 & 69.69 & 50.66 & 53.95 & 61.62  \\
    top-10 & 0.06 &  \cellcolor{col3}68.57 & \cellcolor{col11}78.01 & \cellcolor{col33}59.12 & 69.86 & 50.75 & 54.11 & 61.75 \\
    top-11 & 0.07 & \cellcolor{col3}68.57 & \cellcolor{col11}78.01 & \cellcolor{col33}59.12 & 69.85 & 50.65 & 54.25 & 61.73 \\
    \midrule
    \rowcolor{lightlightgrey}FFT \citep{liu2025few} & -- & \cellcolor{col3}66.29 & \cellcolor{col11}77.01 & \cellcolor{col33}55.56 & 68.25 & 47.88 & 48.44 & 57.66  \\
    \bottomrule
    \end{tabular}}
    \label{tab:layer_selection_dinov2}
\end{table}}

\begin{itemize}
    \item 
    {\bf Checkpoint selection.} 
    We fully finetune a DINOv2 model for 150 epochs and compare checkpoints selected by different methods: our gF1 score, the arithmetic mean of accuracies on the ID training and retrieved data, accuracy on the retrieved data and ID training data only, and no selection (i.e., using the final checkpoint). As shown in Fig.~\ref{fig:ckpt_selection_dinov2}, our gF1 score selects a checkpoint that generalizes robustly to both ID and OOD test data.

    \item 
    {\bf Layer selection.} 
    \Cref{tab:layer_selection_dinov2} reports the results of partially finetuning (PFT) the top-$k$ layers of DINOv2. It reveals a clear trade-off: increasing the number of fine-tuned layers enhances ID accuracy but substantially degrades OOD performance.
    Our gF1 score selects the top-1 layer for PFT, achieving a favorable balance between ID and OOD performance and demonstrating its effectiveness in guiding layer selection for robust adaptation.
    
\end{itemize}

\begin{figure}[t]
  \centering
  \begin{minipage}{0.70\textwidth}
    \includegraphics[width=\linewidth]{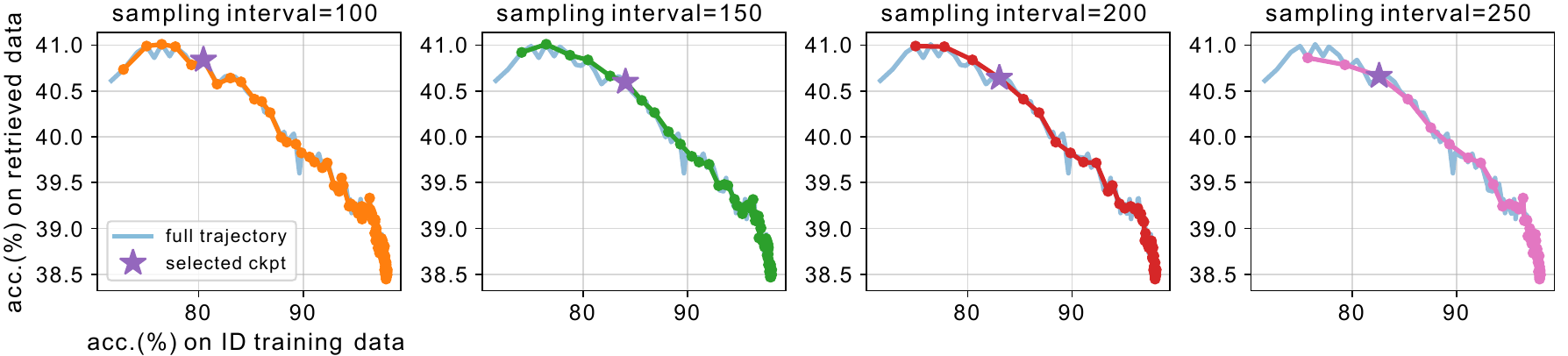}
    \end{minipage}
    \hfill
    \begin{minipage}{0.29\textwidth}
    {
        \setlength{\tabcolsep}{0.7em}
    \scalebox{0.6}{
    \begin{tabular}{lccc}
    \toprule
    \makecell{sampling \\ intervals} & ID test & OOD test & \cellcolor{col3}avg. \\
    \midrule
    100  & 72.41 & 61.05 & \cellcolor{col3}66.73 \\
    150  & 72.55 & 61.03 & \cellcolor{col3}66.79 \\
    200  & 72.74 & 60.88 & \cellcolor{col3}66.81 \\
    250  & 72.58 & 60.94 & \cellcolor{col3}66.76 \\
    \midrule
    \rowcolor{lightlightgrey}w/o val & 72.32 & 59.33 & \cellcolor{col3}65.83 \\
    \bottomrule
    \end{tabular}}}
    \end{minipage}
    \vspace{-3mm}
  \caption{\small
    {\bf Robustness of gF1-based checkpoint selection to checkpoint sampling interval.}
    \textbf{Left}: We finetune CLIP ViT-B/16 on the few-shot training data for 50 epochs and construct checkpoint sets by sampling the fixed training trajectory at intervals of 100, 150, 200, and 250 iterations. 
    Each subplot shows the full training trajectory and the corresponding sampled checkpoints, with the star marking the checkpoint selected by gF1. 
    \textbf{Right}:
    We report the ID and OOD test accuracy, as well as their average, of the checkpoints selected by gF1 under different checkpoint sampling intervals. 
    The selected checkpoints achieve highly consistent performance across all intervals and consistently outperform the validation-free strategy, 
    demonstrating the robustness of gF1 to variations in the checkpoint set.
    }
    \label{fig:sampling_sensitivity}
\end{figure}

{
    \setlength{\tabcolsep}{0.5em}
\begin{table}[t]
  \centering
  \caption{\small
  {\bf Robustness of gF1-based checkpoint selection to training length.}
    We compare gF1-based checkpoint selection with no checkpoint selection under different maximum training lengths for few-shot finetune. gF1 consistently improves both ID and OOD test accuracy across 20, 50, and 100 epochs, demonstrating its robustness to variations in training lengths.}
    \vspace{-3mm}
    \scalebox{0.85}{
    \begin{tabular}{llllllllll}
    \toprule
        \multirow{2}[2]{*}{\makecell{validation \\ method}}  & \multicolumn{3}{c}{epoch=20} & \multicolumn{3}{c}{epoch=50} & \multicolumn{3}{c}{epoch=100} \\
    \cmidrule(lr){2-4}
    \cmidrule(lr){5-7}
    \cmidrule(lr){8-10}
     & ID test & OOD test & \cellcolor{col3}avg.  & ID test & OOD test & \cellcolor{col3}avg.  & ID test & OOD test & \cellcolor{col3}avg. \\
    \midrule
    validation-free & 72.42 & 60.12 & \cellcolor{col3}66.27 & 72.32 & 59.33 & \cellcolor{col3}65.83 & 72.58 & 58.61 & \cellcolor{col3}65.60 \\
    gF1 score & 72.83 & 61.03 & \cellcolor{col3}66.93$^{\textcolor{Green}{+0.66}}$ & 72.58 & 60.94 & \cellcolor{col3}66.76$^{\textcolor{Green}{+0.93}}$ & 72.80  & 60.86 & \cellcolor{col3}66.83$^{\textcolor{Green}{+1.23}}$ \\
    \bottomrule
    \end{tabular}}
  \label{tab:training_length_sensitivity}
\end{table}}

{
\setlength{\tabcolsep}{0.55em}
\begin{table}[t]
  \centering
  \caption{\small
  {\bf Sensitivity of gF1 to reference baselines for layer selection.}
  The original reference baselines, which correspond to the red dashed line in Fig.~\ref{fig:layer-selection}, are highlighted in bold.
  \textbf{Left}: Average test performance over ID and OOD test data of the checkpoints selected by gF1 under different minimum accuracy values.
  \textbf{Right}: Corresponding number of selected layers. 
  The results show that gF1 remains robust to a broad range of minimum accuracy values, with stable test performance and layer selection.
  Moreover, selected checkpoints remarkably outperform full finetuning, which achieves an average test accuracy of 65.83\% (refer to Fig.~\ref{fig:layer-selection}).
  }
  \vspace{-3mm}
  \begin{minipage}{0.48\linewidth}
      \scalebox{0.65}{
  \renewcommand{\arraystretch}{1.15}
    \begin{tabular}{c|l|ccccccc}
    \hline   
    \multicolumn{2}{c|}{} & \multicolumn{7}{c}{min retrieved acc.} \\
    \cline{3-9}   
    \multicolumn{2}{c|}{}  & 38.84   & 38.34     & 37.84   & \textbf{37.34} & 36.84  & 36.34    & 35.84 \\
    \hline
    \multirow{7}[2]{*}{\rotatebox{90}{min ID training acc.}} & 76.09   & 67.67 & 67.66 & 67.84 & 67.99 & 67.78 & 67.78 & 67.78 \\
          & 74.59     & 67.58 & 67.67 & 67.67 & 67.84 & 67.99 & 67.78 & 67.78 \\
          & 73.09   & 67.58 & 67.58 & 67.67 & 67.67 & 67.84 & 67.99 & 67.78 \\
          & \textbf{71.59} & 67.58 & 67.58 & 67.58 & 67.67 & 67.84 & 67.84 & 67.78 \\
          & 70.09  & 67.26 & 67.58 & 67.58 & 67.58 & 67.67 & 67.84 & 67.84 \\
          & 68.59    & 67.26 & 67.26 & 67.58 & 67.58 & 67.67 & 67.67 & 67.84 \\
          & 67.09  & 67.26 & 67.26 & 67.58 & 67.58 & 67.58 & 67.67 & 67.84 \\
    \hline
    \end{tabular}}
  \end{minipage}
  \hfill
  \begin{minipage}{0.48\linewidth}
      \scalebox{0.65}{
  \renewcommand{\arraystretch}{1.15}
    \begin{tabular}{c|l|ccccccc}
    \hline   
    \multicolumn{2}{c|}{} & \multicolumn{7}{c}{min retrieved acc.} \\
    \cline{3-9}   
    \multicolumn{2}{c|}{}  & 38.84   & 38.34     & 37.84   & \textbf{37.34} & 36.84  & 36.34    & 35.84 \\
    \hline
    \multirow{7}[2]{*}{\rotatebox{90}{min ID training acc.}}
    & 76.09   & 4     & 5     & 4     & 3     & 3     & 3     & 3 \\
    & 74.59   & 4     & 4     & 5     & 4     & 3     & 3     & 3 \\
    & 73.09     & 4     & 4     & 4     & 5     & 4     & 3     & 3 \\
    & \textbf{71.59} & 4     & 4     & 4     & 4     & 4     & 4     & 3 \\
    & 70.09  & 5     & 4     & 4     & 4     & 4     & 4     & 4 \\
    & 68.59    & 5     & 5     & 4     & 4     & 4     & 4     & 4 \\
    & 67.09  & 5     & 5     & 4     & 4     & 4     & 4     & 4 \\

    \hline
    \end{tabular}}
  \end{minipage}
  \label{tab:layer_selection_sensitivity}
\end{table}
}

{\bf Robustness analysis.}
In addition to the analysis of gF1's robustness to different reference baselines shown in Fig.~\ref{fig:layer_selection_sensitivity}, 
we present more analyses regarding different aspects.
\begin{itemize}
    \item {\bf Robustness to checkpoint sampling intervals.} 
    \Cref{fig:sampling_sensitivity} evaluates gF1-selected checkpoints from those sampled with different intervals from a training course of 50 epochs. 
    The gF1 selects checkpoints with highly consistent ID and OOD performance across different sampling intervals, consistently outperforming the validation-free strategy.

    \item {\bf Robustness to training length.}
    \Cref{tab:training_length_sensitivity} evaluates the effect of training lengths.
    With different pre-defined training epochs, we sample checkpoints after each training epoch and apply gF1 to select a single checkpoint.
    The results demonstrate that gF1 consistently improves both ID and OOD accuracy over the validation-free strategy, which defaults to the last checkpoint.

    \item {\bf Robustness to perturbations of reference baselines.} 
    Supplementing \Cref{fig:layer_selection_sensitivity},
    we make \Cref{tab:layer_selection_sensitivity} that lists exact accuracies of gF1-selected checkpoints when gF1 is calculated with different reference baselines (Eq.~\ref{eq:gF1}), as well as the epoch numbers for these checkpoints.    
    The selected checkpoints are from neighboring layer depths (top-3 to top-5), 
    maintain stable test performance, 
    and consistently outperform the full finetuning method \citep{liu2025few}.
\end{itemize}

\begin{figure}[t]  
    \begin{minipage}{0.33\textwidth}
    \centering
    \vspace{5mm}
    \includegraphics[width=0.93\linewidth]{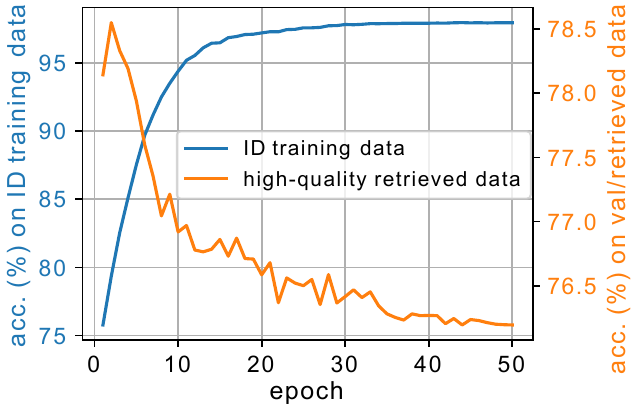}
    \end{minipage}
    \hfill
    \begin{minipage}{0.66\textwidth}
    \centering
        \includegraphics[width=\linewidth]{figures/delimma_v8.pdf}
    \end{minipage}
    \vspace{-3mm}
    \caption{
        {\bf Paradox remains pronounced with high-quality retrieved data for validation.}
        \textbf{Left:} We plot the accuracy of high-quality retrieved data used in \Cref{tab:gF1_sensitive} throughout few-shot finetuning together with the ID training accuracy.
        The mid and right panels copy Fig.~\ref{fig:dilemma}.
        Despite the higher quality of the retrieved data, the paradox observed in Fig.~\ref{fig:dilemma} remains.
        This paradox lays the foundation for gF1 to work effectively.
        }
        \label{fig:delimma_suppl}
\end{figure}

{\bf Paradox in Fig.~\ref{fig:dilemma} remains when using high-quality retrieved data for validation.}
One may wonder if using high-quality retrieved data change the paradox presented in Fig.~\ref{fig:dilemma}.
To address this,
we use high-quality retrieved data as those that have the top-50 highest feature similarities to the labeled training samples (ref. ``high-quality'' in Table~\ref{tab:gF1_sensitive}).
Fig.~\ref{fig:delimma_suppl}-left depicts the checkpoints' training accuracy and accuracy on this retrieved dataset,
alongside Fig.~\ref{fig:dilemma} copied on the right.
Clearly, 
the paradox remains, showing a foundation for gF1 to work effectively.
This also re-confirms the effectiveness of gF1 when using different levels of retrieved data quality as shown in Table~\ref{tab:gF1_sensitive}.

\section{Details of Compared Methods}
\label{sec:compared_methods}

To evaluate the performance stability of both baselines and the proposed method, we conduct three runs with different 16-shot training data splits and compute the mean accuracy and standard deviations.
Results in Table~\ref{tab:random_seed} demonstrate that our method exhibits low variance and hence better stability.
We describe the compared methods below, categorized into four groups.
\begin{itemize}
    \item \textbf{Prompt Tuning} methods adapt a VLM through learnable prompt tokens without adapting the entire model. Among them, CoOp~\citep{zhou2022learning} and CoCoOp~\citep{zhou2022conditional} optimize text prompt tokens for the language branch, while VPT~\citep{jia2022visual} introduces learnable prompts for visual inputs. MaPLe~\citep{khattak2023maple} simultaneously learns prompts for both visual and textual modalities. 
    ProText~\citep{Khattak2024ProText} leverages large language models to learn text prompts without visual supervision.
    Promise~\citep{liang2026prompt} introduces a meta-learning framework to enhance the robustness of VLM against variations in natural language prompts.
    
    \item \textbf{Adapter Learning} methods adapt VLM by introducing lightweight trainable modules while keeping the backbone frozen. Tip-Adapter~\citep{zhang2022tip} implements a key-value cache mechanism derived from few-shot training examples. CLIP-Adapter~\citep{gao2023clip} introduces an adapter consisting of two linear layers. TaskRes~\citep{yu2023task} learns a target task classifier by optimizing a set of task-specific residual parameters. DMN-ZS~\citep{zhang2024dual} proposes dual memory networks to preserve knowledge of few-shot training data and testing data. LoRA~\citep{hu2022lora} adapts the large language model by incorporating learnable low-rank matrices into model weights. Following the implementation of~\citep{zanella2024low}, we apply LoRA specifically to the vision encoder of VLM. 

    \item \textbf{Linear Probing} methods adapt pretrained VLM by training a linear classifier on the frozen visual encoder. LP w/ WiSE-FT~\citep{wortsman2022robust} employs weight-space ensembling, combining classifier weights with the pretrained model. We use an interpolation factor of 0.5 in our experiment. CMLP~\citep{lin2023multimodality} leverages multimodal examples as additional training examples. LP w/ LCA~\citep{shi2024lca} introduces WordNet class hierarchy as the soft label. CLAP~\citep{clap24} introduces class-adaptive penalty terms during classifier training.
    
    \item \textbf{Finetuning} methods adapt VLM by updating all parameters for downstream tasks. Standard finetuning method only finetunes the visual encoder with cross entropy loss~\citep{liu2025few}, while FLYP finetunes both visual and textual encoders with contrastive loss~\citep{goyal2023finetune}. SWAT~\citep{liu2025few} proposes a two-stage method with retrieval augmentation for few-shot learning.
\end{itemize}

{
\setlength{\tabcolsep}{0.5em} 
\begin{table*}[t]
  \centering
  \caption{\small {\bf Averaged accuracy with standard deviations.} For each method, we run three times using different few-shot training data and report the average accuracy and standard deviation on ImageNet (as ID test set) and its variants (as OOD test sets). Our proposed methods demonstrate superior accuracy and stability.
  }
    \small
      \vspace{-3mm}
    \scalebox{0.8}{
    \begin{tabular}{lllcccccc}
    \toprule
    & & & ID test & \multicolumn{5}{c}{OOD test} \\
    \cmidrule(lr){4-4}
    \cmidrule(lr){5-9}
    & methods & validation  & IN & IN-V2 & IN-S & IN-A & IN-R & avg. \\
    \midrule
    \multirow{4}{*}{\makecell{Prompt \\ Tuning}}      & CoOp \scriptsize{IJCV'22}  & {\color{gray}ID test acc.} & 
    71.74{\color{gray}$\pm$0.16} & 64.57{\color{gray}$\pm$0.22}  & 47.54{\color{gray}$\pm$0.31} & 50.40{\color{gray}$\pm$0.25} & 75.58{\color{gray}$\pm$0.36}  & 59.52{\color{gray}$\pm$0.28}  \\
          & MaPLe \scriptsize{CVPR'23} & {\color{gray}ID test acc.} & 
          71.05{\color{gray}$\pm$0.17}  & 64.27{\color{gray}$\pm$0.33}  & 48.74{\color{gray}$\pm$0.45}  & 50.72{\color{gray}$\pm$0.20}  & 77.22{\color{gray}$\pm$0.34}  & 60.24{\color{gray}$\pm$0.29}  \\
          & VPT{\scriptsize shallow} \scriptsize{ECCV'22} & {\color{gray}ID test acc.} & 
          72.61{\color{gray}$\pm$0.20}  & 65.51{\color{gray}$\pm$0.30}  & 49.08{\color{gray}$\pm$0.21}  & 49.72{\color{gray}$\pm$0.47}  & 78.51{\color{gray}$\pm$0.13}  & 60.70{\color{gray}$\pm$0.23}  \\
          & VPT{\scriptsize deep} \scriptsize{ECCV'22} & {\color{gray}ID test acc.} &
          72.63{\color{gray}$\pm$0.14}  & 65.14{\color{gray}$\pm$0.23}  & 47.67{\color{gray}$\pm$0.23}  & 47.28{\color{gray}$\pm$0.50}  & 76.59{\color{gray}$\pm$0.16}  & 59.17{\color{gray}$\pm$0.26}  \\
    \midrule

    \multirow{5}{*}{\makecell{Adapter \\ Learning}} & Tip-Adapter \scriptsize{ECCV'22} & {\color{gray}ID test acc.} & 
    70.53{\color{gray}$\pm$0.03}  & 63.26{\color{gray}$\pm$0.09}  & 48.68{\color{gray}$\pm$0.06}  & \underline{50.90}{\color{gray}$\pm$0.09}  & 77.80{\color{gray}$\pm$0.07}  & 60.16{\color{gray}$\pm$0.04}  \\
          & Tip-Adapter-F \scriptsize{ECCV'22} & {\color{gray}ID test acc.} &
          73.34{\color{gray}$\pm$0.14}  & 65.32{\color{gray}$\pm$0.19}  & 47.72{\color{gray}$\pm$0.33}  & 48.91{\color{gray}$\pm$0.27}  & 76.63{\color{gray}$\pm$0.29}  & 59.79{\color{gray}$\pm$0.16}  \\
          & CLIP-Adapter \scriptsize{IJCV'23} & {\color{gray}ID test acc.} & 
          71.48{\color{gray}$\pm$0.14}  & 63.88{\color{gray}$\pm$0.15}  & 46.49{\color{gray}$\pm$0.20}  & 47.65{\color{gray}$\pm$0.58}  & 73.78{\color{gray}$\pm$0.36}  & 57.95{\color{gray}$\pm$0.25}  \\
          & TaskRes \scriptsize{CVPR'23} & {\color{gray}ID test acc.} & 
          73.42{\color{gray}$\pm$0.10}  & 65.27{\color{gray}$\pm$0.19}  & 48.08{\color{gray}$\pm$0.05}  & 49.27{\color{gray}$\pm$0.70}  & 76.93{\color{gray}$\pm$0.29}  & 59.89{\color{gray}$\pm$0.16}  \\
          & LoRA \scriptsize{ICLR'22} & {\color{gray}ID test acc.} &
          73.01{\color{gray}$\pm$0.12}  & 65.56{\color{gray}$\pm$0.22}  & 48.59{\color{gray}$\pm$0.20}  & 50.51{\color{gray}$\pm$0.42}  & 77.85{\color{gray}$\pm$0.24}  & 60.63{\color{gray}$\pm$0.09}  \\
    \midrule

    \multirow{5}{*}{\makecell{Linear \\ Probing}} & LP\scriptsize{ICML'21}   & {\color{gray}ID test acc.} & 
    73.54{\color{gray}$\pm$0.20}  & 65.48{\color{gray}$\pm$0.24}  & 48.78{\color{gray}$\pm$0.25}  & 49.59{\color{gray}$\pm$0.21}  & 77.65{\color{gray}$\pm$0.45}  & 60.37{\color{gray}$\pm$0.16}  \\
          & LP w/ WiSE-FT \scriptsize{CVPR'22} & {\color{gray}ID test acc.} & 
          71.93{\color{gray}$\pm$0.04}  & 64.66{\color{gray}$\pm$0.22}  & 49.44{\color{gray}$\pm$0.01}  & 50.62{\color{gray}$\pm$0.07}  & 78.33{\color{gray}$\pm$0.03}  & 60.76{\color{gray}$\pm$0.04}  \\
          & CMLP \scriptsize{CVPR'23} & {\color{gray}ID test acc.} & 
          73.38{\color{gray}$\pm$0.07}  & 65.14{\color{gray}$\pm$0.30}  & 48.52{\color{gray}$\pm$0.33}  & 49.10{\color{gray}$\pm$0.48}  & 77.42{\color{gray}$\pm$0.17}  & 60.05{\color{gray}$\pm$0.11}  \\
          & LP w/ LCA \scriptsize{ICML'24} & {\color{gray}ID test acc.} & 
          69.34{\color{gray}$\pm$0.07}  & 62.53{\color{gray}$\pm$0.11}  & 48.52{\color{gray}$\pm$0.02}  & 49.47{\color{gray}$\pm$0.19}  & 77.76{\color{gray}$\pm$0.05}  & 59.57{\color{gray}$\pm$0.05}  \\
          & CLAP \scriptsize{CVPR'24} & val-free & 72.86{\color{gray}$\pm$0.13}  & 65.06{\color{gray}$\pm$0.06}  & 47.79{\color{gray}$\pm$0.19}  & 49.58{\color{gray}$\pm$0.22}  & 77.09{\color{gray}$\pm$0.24}  & 59.88{\color{gray}$\pm$0.08}  \\
    \midrule

    \multirow{4}{*}{Finetuning} & FLYP \scriptsize{CVPR'23} & {\color{gray}ID test acc.} & 
    \underline{74.14}{\color{gray}$\pm$0.16}  & 65.50{\color{gray}$\pm$0.05}  & 47.29{\color{gray}$\pm$0.09}  & 45.43{\color{gray}$\pm$0.62}  & 74.82{\color{gray}$\pm$0.34}  & 58.26{\color{gray}$\pm$0.13}  \\
          & FFT \scriptsize{CVPR'25} & val-free & 72.37{\color{gray}$\pm$0.10}  & 64.85{\color{gray}$\pm$0.25}  & 48.09{\color{gray}$\pm$0.30}  & 48.97{\color{gray}$\pm$0.33}  & 75.67{\color{gray}$\pm$0.50}  & 59.39{\color{gray}$\pm$0.20}  \\
          & SWAT \scriptsize{CVPR'25} & val-free & 
          73.56{\color{gray}$\pm$0.07}  & 
          \underline{66.57}{\color{gray}$\pm$0.15}  & 
          \textbf{54.89}{\color{gray}$\pm$0.09}  & 50.85{\color{gray}$\pm$0.23}  & 
          \textbf{81.18}{\color{gray}$\pm$0.07}  & \underline{63.37}{\color{gray}$\pm$0.03}  \\
          & \textbf{VEST} \scriptsize{ours} & gF1 score & \textbf{75.43}{\color{gray}$\pm$0.11}  & 
          \textbf{68.45}{\color{gray}$\pm$0.01}  & \underline{54.57}{\color{gray}$\pm$0.02}  & 
          \textbf{52.94}{\color{gray}$\pm$0.02}  & \underline{81.08}{\color{gray}$\pm$0.08}  & 
          \textbf{64.26}{\color{gray}$\pm$0.02}  \\

    \bottomrule
    \end{tabular}}%
  \label{tab:random_seed}%
  \vspace{-1mm}
\end{table*}}

\begin{table}[t]
  \centering
  \begin{minipage}{0.47\textwidth}
    \caption{\small
    {\bf gF1 enables VEST to achieve strong performance on Living17 and Entity30.} 
    We report the top-1 accuracy on both ID and OOD test data. 
    We compare our methods against few-shot finetuning (FFT) and the state-of-the-art FSR approach, SWAT. By enabling reliable checkpoint and layer selection via gF1 score, our VEST achieves the best performance among all compared methods.
    }
    \vspace{-3mm}
    \setlength{\tabcolsep}{0.7em}    
    \scalebox{0.7}{
    \begin{tabular}{llcccc}
    \toprule
    \multirow{2}[4]{*}{method} & \multirow{2}[4]{*}{validation} & \multicolumn{2}{c}{Living17} & \multicolumn{2}{c}{Entity30} \\
    \cmidrule(lr){3-4}
    \cmidrule(lr){5-6}
    && \cellcolor{col11}ID    & \cellcolor{col33}OOD   & \cellcolor{col11}ID    & \cellcolor{col33}OOD \\
    \midrule
    zero-shot & -- & \cellcolor{col11}89.77 & \cellcolor{col33}93.71 & \cellcolor{col11}70.17 & \cellcolor{col33}70.25 \\
    \midrule
    FFT  & val-free & \cellcolor{col11}92.65 & \cellcolor{col33}\underline{94.00}  & \cellcolor{col11}81.93 & \cellcolor{col33}\underline{75.60} \\
    FFT  & gF1 & \cellcolor{col11}92.53 & \cellcolor{col33}93.82 & \cellcolor{col11}81.88 & \cellcolor{col33}75.57 \\
    SWAT & val-free & \cellcolor{col11}\textbf{96.88} & \cellcolor{col33}89.12 & \cellcolor{col11}\textbf{92.83} & \cellcolor{col33}69.40 \\
    PFT \scriptsize{ours} & gF1 & \cellcolor{col11}92.12 & \cellcolor{col33}\textbf{94.18} &\cellcolor{col11}82.53 & \cellcolor{col33}\textbf{76.18} \\
    VEST \scriptsize{ours} & gF1 & \cellcolor{col11}\underline{96.77} & \cellcolor{col33}92.12 & \cellcolor{col11}\underline{92.68} & \cellcolor{col33}72.40 \\
    \bottomrule
    \end{tabular}}
  \label{tab:breeds}
  \end{minipage}
    \hspace{\fill}
    \begin{minipage}{0.49\textwidth}
    \caption{\small
    {\bf gF1 enables effective layer selection for PFT on fMoW.}
    We report the top-1 accuracy on both ID and OOD test data, along with the average OOD performance. 
    Notably, consistent with prior work~\citep{parashar2024neglected, liu2025few}, a detrimental domain gap exists between retrieved and few-shot ID satellite data.
    Yet, gF1-selected PFT yields consistent improvements on both ID and OOD data.
    }
    \vspace{-3mm}
    \setlength{\tabcolsep}{0.5em}    
    \scalebox{0.8}{
    \begin{tabular}{llcccc}
    \toprule
    \multirow{3}[3]{*}{method} & \multirow{3}[4]{*}{validation}  & ID    & \multicolumn{3}{c}{OOD} \\
    \cmidrule(lr){3-3}    
    \cmidrule(lr){4-6}
     && \cellcolor{col11}Americas & \cellcolor{col33}avg.  & Africa & Europe \\
    \midrule
    zero-shot & -- & \cellcolor{col11}18.38 & \cellcolor{col33}17.19 & 15.66 & 18.72 \\
    \midrule
    FFT & val-free & \cellcolor{col11}39.59 & \cellcolor{col33}28.89 & 23.08 & 34.70 \\
    FFT & gF1 & \cellcolor{col11}39.27 & \cellcolor{col33}\underline{29.58} & \underline{24.20} & \underline{34.96} \\
    SWAT & val-free & \cellcolor{col11}38.80  & \cellcolor{col33}25.87 & 21.96 & 29.79 \\
    PFT \scriptsize{ours} & gF1 & \cellcolor{col11}\underline{39.72} & \cellcolor{col33}\textbf{30.19} & \textbf{24.85} & \textbf{35.53} \\
    VEST \scriptsize{ours} & gF1 & \cellcolor{col11}\textbf{40.32} & \cellcolor{col33}27.84 & 22.39 & 33.28 \\
    \bottomrule
    \end{tabular}}
    \label{tab:fMow}
    \end{minipage}
\end{table}

{
\setlength{\tabcolsep}{0.5em} 
\begin{table}[t]
    \centering
  \caption{\small
  {\bf gF1 enables VEST to achieve strong performance on DomainNet dataset.}
Using gF1 for checkpoint selection improves performance over the final checkpoint of few-shot finetuning (FFT), a default validation-free configuration. 
Furthermore, gF1 enables layer selection for partial finetuning (PFT), leading to further gains.
With the proposed validation strategy, our VEST significantly outperforms SWAT on both ID and OOD test data.
  }
  \vspace{-3mm}
  \scalebox{0.85}{
    \begin{tabular}{llccccccc}
    \toprule
    \multirow{2}[3]{*}{method} & \multirow{2}[3]{*}{validation} & ID & \multicolumn{6}{c}{OOD} \\
    \cmidrule(lr){3-3}
    \cmidrule(lr){4-9}
    &       & \cellcolor{col11}Sketch & \cellcolor{col33}avg. & Real & Quickdraw & Painting & Infograph & Clipart \\
    \midrule
    zero-shot & --    & \cellcolor{col11}62.83 & \cellcolor{col33}55.80  & 83.23 & 13.84 & 65.12 & 45.61 & 70.19 \\
    \midrule
    FFT   & val-free & \cellcolor{col11}70.17 & \cellcolor{col33}58.83 & 83.70  & 17.52 & 67.67 & 51.02 & 74.23 \\
    FFT   & gF1   & \cellcolor{col11}\underline{70.34} & \cellcolor{col33}59.34 & 84.37 & 18.06 & 68.50  & 50.85 & 74.93 \\
    SWAT  & val-free & \cellcolor{col11}69.95 & \cellcolor{col33}59.48 & 83.17 & \underline{18.71} & 68.30  & \underline{53.06} & 74.16 \\
    PFT \scriptsize{ours} & gF1   & \cellcolor{col11}\textbf{70.41} & \cellcolor{col33}\underline{59.50}  & \underline{84.54} & 18.09 & \underline{68.93} & 51.26 & \underline{74.70} \\
    VEST \scriptsize{ours} & gF1   & \cellcolor{col11}70.08 & \cellcolor{col33}\textbf{60.58} & \textbf{84.64} & \textbf{18.80}  & \textbf{69.93} & \textbf{54.13} & \textbf{75.42} \\
    \bottomrule
    \end{tabular}}
  \label{tab:domainnet}
  \vspace{-1mm}
\end{table}}

\section{Results on More Datasets}
\label{sec:other_datasets}

\begin{figure}[t]
    \centering
    \includegraphics[width=0.99\linewidth]{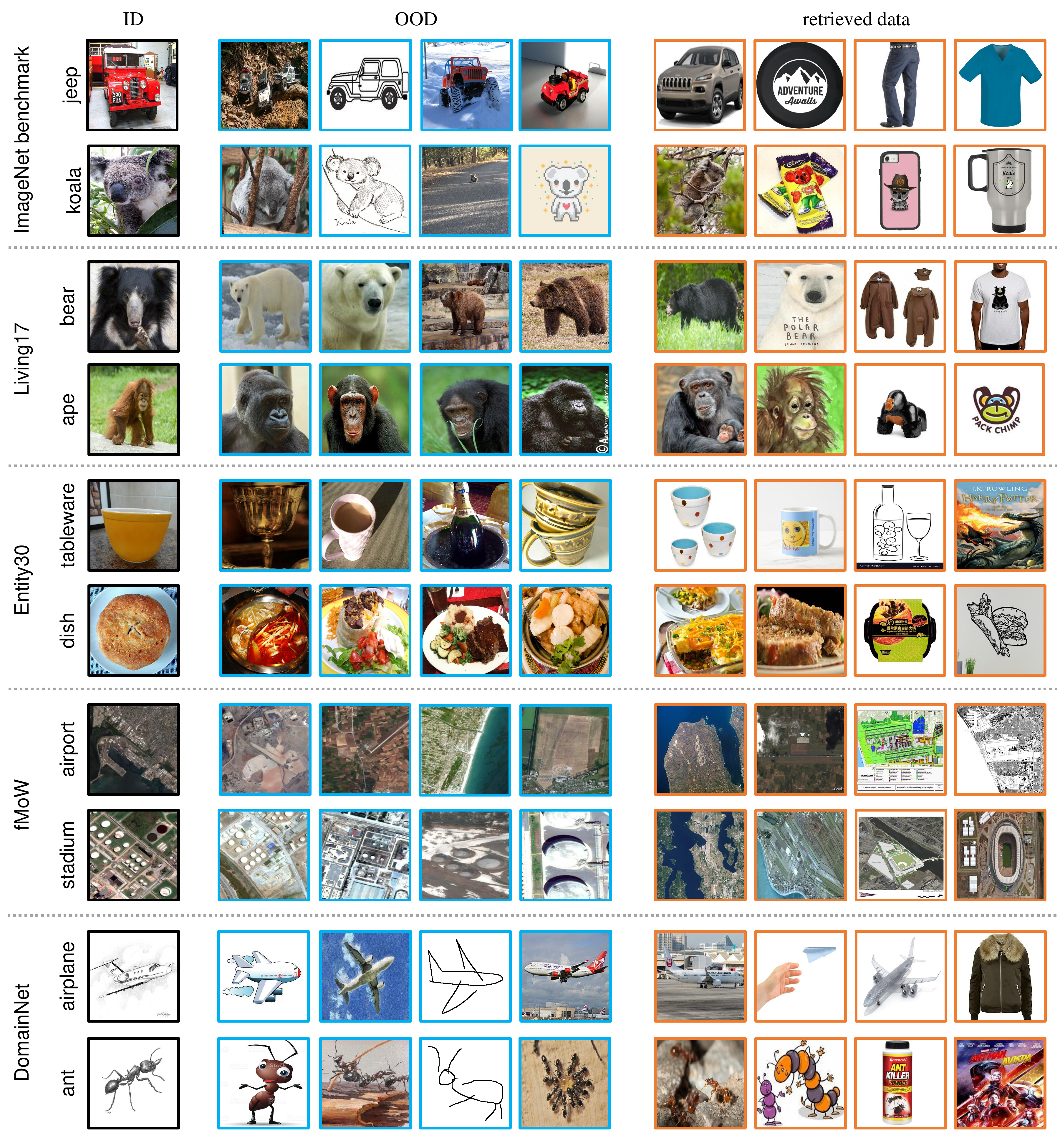}
    \vspace{-3mm}
    \caption{\small
    {\bf Comparing} ID training data, OOD test  data (marked by \textcolor[RGB]{92, 169, 241}{blue border}), and retrieved data from LAION-400M (marked by \textcolor[RGB]{219, 133, 50}{orange border}). To illustrate the distributional characteristics, we present examples of the training and testing images by randomly selecting two classes from each dataset. The comparison reveals significant distribution shifts between ID and OOD data. Notably, although the retrieved data enhances sample diversity, it also introduces noisy images due to linguistic ambiguity in the retrieval process \citep{liu2025few}.
    }
    \label{fig:other_datasets_supp}
\end{figure}

We further evaluate the effectiveness of the proposed validation strategy and framework on more datasets using the CLIP ViT-B/16 model in \Cref{tab:breeds}, \Cref{tab:fMow}, and \Cref{tab:domainnet}. Refer to Fig.~\ref{fig:other_datasets_supp} for the visual examples from these datasets.
We compare our methods with few-shot finetuning (FFT) and the recent FSR approach, SWAT~\citep{liu2025few}. Our gF1 enhances FFT by enabling reliable checkpoint selection and further supports layer selection. Building on this validation strategy, the full framework VEST achieves higher ID and OOD accuracies than SWAT.

\section{Discussions on Limitations and Future Work}
\label{sec:more_discussions}
We note certain limitations that warrant a discussion.
Our approach relies on retrieving task-relevant external data, which may be challenging or even infeasible for highly specialized domains with limited relevant public data. One potential solution is to leverage synthetic images for both validation and training data augmentation. 
We leave it for future work.
Moreover, gF1 assigns equal weights to the two objectives: performance gain on ID training data and performance preservation on retrieved data. However, their relative importance may vary across tasks and domains. 
A promising direction is to develop task-adaptive weighting strategies, for example, by measuring the domain gap between ID and retrieved data using MMD~\citep{gretton2012kernel} or CORAL~\citep{dcoral} and using this measure to adapt the relative weights of the two objectives.

\section{Open-Source Code}
\label{sec:code}

We include our code in the supplementary material, comprising the following components.

{\bf Requirements.} Running the code requires Python, PyTorch, and other common packages. We list all dependencies in the \textit{requirements.txt} file for convenient installation. Additionally, we include detailed environment setup instructions in the \textit{README.md} file. We used Python 3.10.16 and PyTorch 2.0.0 for experiments.

{\bf License.} We will release our code under the MIT license.

{\bf Instructions.} We provide detailed instructions for reproducing our experiments in the following markdown files:
\begin{itemize}
    \item \textit{DATASETS.md} provides detailed steps for installing the datasets used in our experiments, including ImageNet-based OOD benchmark, few-shot splits, and retrieval dataset.
    \item \textit{README.md} provides detailed guidelines on how to set up the environment and run the demo of the proposed methods.
\end{itemize}

{\bf Demo.} The code includes three demos to demonstrate the results of our proposed methods in Table~\ref{tab:components} and Table~\ref{tab:baseline_dinov2}. In the demonstration, we finetune CLIP ViT-B/16 \citep{cherti2023reproducible} and DINOv2 ViT-B/14 \citep{oquab2024dinov} models for 16-shot learning using ImageNet (IN) as the ID dataset and evaluate the methods both on ImageNet and its four OOD variants, including ImageNet-V2 (IN-V2)~\citep{recht2019imagenet}, 
ImageNet-Sketch (IN-S)~\citep{wang2019learning}, 
ImageNet-Adversarial (IN-A)~\citep{hendrycks2021natural}, 
and ImageNet-Rendition (IN-R)~\citep{hendrycks2021many}. The details of the demos are as follows:
\begin{itemize}
    \item \textit{PFT\_demo.ipynb} contains the implementation of \textit{Partial Finetuning} using CLIP model.
    \item \textit{VEST\_demo.ipynb} contains the implementation of \textit{Validation-Enabled Stage-wise Tuning} using CLIP model. 
    \item \textit{VEST\_dinov2\_demo.ipynb} contains the implementation of \textit{Validation-Enabled Stage-wise Tuning} using DINOv2. 
    
\end{itemize}

\end{document}